%% file: neurips_2026.tex
\documentclass{article}

\PassOptionsToPackage{numbers, compress}{natbib}
\usepackage[preprint]{neurips_2026}

\usepackage[utf8]{inputenc} 
\usepackage[T1]{fontenc}    
\usepackage{hyperref}       
\usepackage{url}            
\usepackage{booktabs}       
\usepackage{amsfonts}       
\usepackage{nicefrac}       
\usepackage{microtype}      
\usepackage{xcolor}         

\setcitestyle{square}
\usepackage{amsmath}
\usepackage{graphicx}
\usepackage[capitalise]{cleveref}
\usepackage{multirow}
\usepackage{amssymb}

\title{DecMem: Towards Minute-Long Consistent World Generation with Decoupled Memory}

\author{%
  Zhenhao Yang$^{1*}$\;, Xiaoshi Wu$^2$, Zhengyao Lv$^1$, Xiaoyu Shi$^{2\dagger}$ \\ \textbf{Xintao Wang$^2$, Pengfei Wan$^2$, Kun Gai$^2$, Kwan-Yee K. Wong$^{1\dagger}$}\\
  $^1$The University of Hong Kong, $^2$Kling Team, Kuaishou Technology\\
}

\begin{document}

\maketitle

\renewcommand{\thefootnote}{}
\footnotetext{\parbox[t]{\linewidth}{\raggedright
*Work done during an internship at Kling Team, Kuaishou Tech.\\
$^{\dag}$Corresponding Author.}}

\input{sec/0_abstract}

\input{sec/1_introduction}
\input{sec/2_relatedwork}
\input{sec/3_method}
\input{sec/4_experiment}
\input{sec/5_conclusion}

\bibliographystyle{plainnat}
\bibliography{main}

\newpage
\appendix
\input{sec/6_appendix}





\end{document}

%% file: sec/0_abstract.tex
\begin{abstract}
  Recent advances in video generative models have promoted rapid progress in controllable world models. However, maintaining fine-grained spatio-temporal consistency under long-horizon reasoning remains a key challenge. In this work, we move beyond explicit 3D memory and coarse frame-level implicit modeling, and propose a fine-grained, learnable, and scalable memory for consistent world generation. We first identify two fundamental limitations of na\"ive learnable memory architectures in long-horizon extrapolation, namely computational inefficiency and attention dispersion. Through a systematic analysis of attention dispersion, we propose DecMem, a decoupled memory architecture that employs Sparse Global Memory for efficient fine-grained access to global history and Anchored Local Memory for stable and high-quality extrapolation. Extensive experiments demonstrate that DecMem significantly outperforms current state-of-the-art methods. By ensuring precise and efficient long-term memory and achieving superior extrapolation capabilities, DecMem enables minute-level controllable long video generation with high fidelity and consistency. Project page is available at \url{https://jeffreyyzh.github.io/DecMem-Page}
\end{abstract}


%% file: sec/1_introduction.tex
\section{Introduction}
\label{sec:intro}

With the rapid evolution of generative video modeling, leveraging powerful pretrained video generation backbones~\citep{wan2025wan, kong2024hunyuanvideo} to construct world models has become a pivotal research frontier. While recent works have successfully achieved controllable generation through injecting action information~\citep{li2025hunyuan-gamecraft, tang2025hunyuan-gamecraft2, he2025matrix-game-2, yu2025gamefactory, valevski2024GameNgen, mao2025yume-1.5, zhang2025matrix-game,xiang2025pan,ye2025yan}, generating high-quality and consistent long videos remains a formidable challenge. This issue is particularly pronounced in ``revisit'' scenarios, where existing models frequently fail to recall previously generated scenes as inference extends, leading to significant temporal inconsistencies. Fundamentally, building a temporally consistent world model demands flexible and efficient exploitation of long-term memory, rather than being confined to local context mechanisms such as sliding windows~\citep{henschel2025streamingt2v, teng2025magi1, chen2024diffusion_forcing, huang2025self_forcing, chen2025skyreels, yin2025causvid} or their extensions that incorporate attention sinks~\citep{yang2025longlive, liu2025rollingforcing,cui2026lol,shin2025motionstream}.

\begin{figure}[t]
    \centering
    \includegraphics[width=\linewidth]{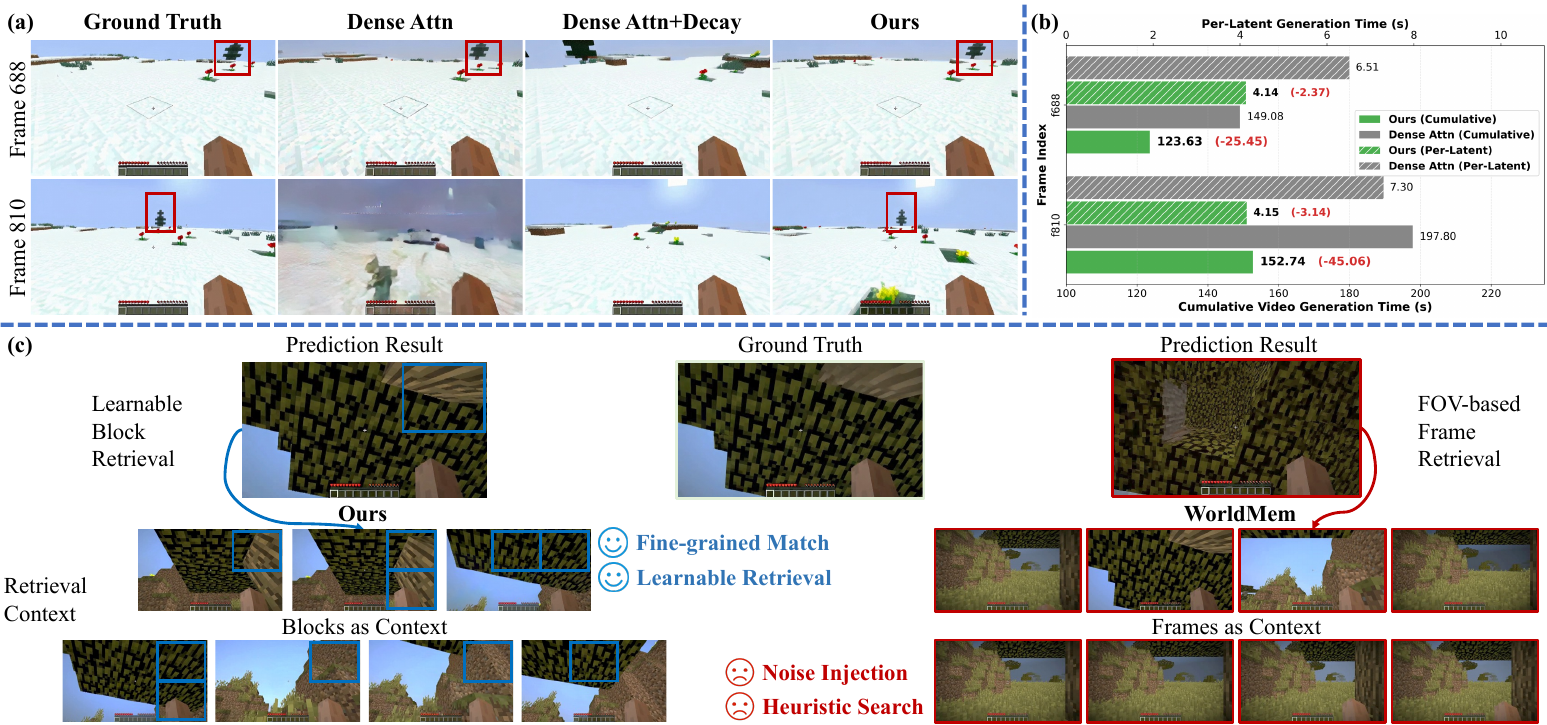}
    \caption{(a) Visual quality and spatio-temporal consistency of different long-horizon extrapolation methods (memory bank initialized with 221 frames). Prior methods fail to jointly preserve fidelity and consistency, while ours breaks this trade-off and sustains fine-grained memory under long rollouts. (b) Generation latency of our method and na\"ive Dense Attention (memory bank initialized with 221 frames). Our sparse block retrieval substantially reduces cost without sacrificing quality. (c) Comparison of our learnable block retrieval against FOV-based frame retrieval (e.g., WorldMem).}
    \label{fig:intro}
\end{figure}


Existing memory mechanisms can be broadly classified into two categories, namely explicit memory and implicit memory. Explicit Memory approaches rely on explicitly constructed 3D representations~\citep{wu2025spmem,huang2025memory_forcing,li2025vmem,zhao2025spatia,duan2026liveworld,wang2026anchorweave,team2025hunyuanworld}. While geometric priors naturally favor spatial consistency, their performance is bounded by the underlying 3D estimator. Maintaining 3D representations incurs additional overheads, and estimation errors accumulated over time can erode long-range consistency. Early implicit memory approaches~\cite{xiao2025worldmem, yu2025contextas, sun2025worldplay} leverage camera poses and field-of-view (FOV) to retrieve relevant frames from a memory bank, thereby expanding their effective context window (see~\cref{fig:intro}(c)). Attention-based implicit memory approaches~\cite{chen2026out, xiang2026viewrope,chen2025vrag}, on the other hand, model inter-frame dependencies implicitly in the attention mechanism. While all these implicit memory approaches advance toward learnable memory, they remain bounded by a frame-level granularity bottleneck. FOV-based approaches rely on heuristic policies that cannot be jointly optimized with the generative objective, whereas attention-based approaches, though end-to-end learnable, treat each frame as an indivisible unit and fail to capture sub-frame spatio-temporal correspondences. 

To overcome the granularity bottleneck of implicit memory while avoiding the fragility of explicit 3D representations, a straightforward design is to let every token perform dense attention over all historical features, thereby achieving the finest-grained and fully learnable long-term memory. However, this simple design suffers from two fundamental limitations, namely attention dispersion and computational inefficiency. As the context grows, we observe a flood of weakly-relevant historical features which dilutes the attention weights allocated to the critical ones. Such an attention dispersion causes severe quality degradation and structural collapse (\cref{fig:intro}(a)).
The per-latent generation latency scales linearly with the sequence length, and the overall generation cost grows rapidly. Such a computational inefficiency severely constrains the scalability towards minute-long video synthesis (\cref{fig:intro}(b)).
While some training-free extrapolation methods~\citep{zhao2025ultravico} alleviate the attention dispersion problem by mechanically down-weighting distant tokens, they do so at the cost of long-range memory loss (\cref{fig:intro}(a)), exposing a fundamental dilemma between short-range fidelity and long-range consistency.

To address the aforementioned limitations, we propose a fine-grained, learnable, and scalable \textbf{dec}oupled \textbf{mem}ory architecture, named DecMem, consisting of two complementary modules. The first module is the Sparse Global Memory (SGM), which performs block-level sparse retrieval over the full history to achieve efficient yet fine-grained long-term memory access. The second module is the Anchored Local Memory (ALM), which anchors attention in recent frames to stabilize the attention distribution. By decoupling global retrieval from local anchoring, DecMem resolves the attention dispersion problem and enables scaling to minute-long video synthesis with strong spatio-temporal consistency.


Our main contributions can be summarized as follows:
\begin{itemize}
\item We systematically reveal the root cause of the limited long-horizon extrapolation capability of na\"ive dense-attention designs and pinpoint the intrinsic limitations of training-free strategies in preserving long-range memory. We then propose a fine-grained, learnable, and scalable memory mechanism for long-video world model.
\item We introduce a novel decoupled memory architecture named DecMem, with Sparse Global Memory for global, efficient, and fine-grained memory access, and Anchored Local Memory for explicit mitigation of attention dispersion under long-horizon inference.
\item Our method consistently surpasses current state-of-the-art baselines, achieving minute-long controllable video generation with strong spatio-temporal consistency and visual quality.
\end{itemize}

%% file: sec/2_relatedwork.tex
\section{Related Works}
\textbf{Interactive World Model.}
Driven by the remarkable success of diffusion methods~\cite{rombach2022stablediffusoon, peebles2023dit} in high-fidelity video generation~\citep{wan2025wan, kong2024hunyuanvideo}, leveraging these generative priors to construct controllable world models has emerged as a pivotal research direction. Early explorations, such as GameNGen~\citep{valevski2024GameNgen} and Matrix~\citep{feng2024thematrix}, primarily utilized discrete keyboard information as control signals, while subsequent works~\citep{genie3, yu2025gamefactory, zhang2025matrix-game, he2025matrix-game-2, li2025hunyuan-gamecraft} incorporated mouse trajectories to enable precise view-dependent interactions. More recently, Hunyuan-GameCraft2~\citep{tang2025hunyuan-gamecraft2} and Yume-1.5~\citep{mao2025yume-1.5} have integrated prompt-based instructions to trigger new events. However, maintaining robust long-term consistency in world simulation remains a key challenge.

\noindent\textbf{Memory Retrieval.}
To achieve spatio-temporal consistency in long-video generation, one line of work~\citep{wu2025spmem,huang2025memory_forcing,li2025vmem,zhao2025spatia,duan2026liveworld,wang2026anchorweave,team2025hunyuanworld} explicitly constructs geometric representations to establish spatial correspondences between a target frame and the historic frames stored in a memory bank. While such explicit memory mechanisms enable precise spatial association, their performance is bounded by the accuracy of the underlying 3D estimator, with estimation errors accumulate as generation extends. To circumvent the fragility of explicit 3D representations, an alternative line of work resorts to implicit memory, for instance, by leveraging camera poses and field-of-view (FOV)~\citep{xiao2025worldmem,sun2025worldplay,yu2025contextas} to retrieve relevant frames. Despite their efficacy, these explicit retrieval mechanisms often ignore the potential of learning-based optimization.

Existing learnable approaches \cite{chen2026out,xiang2026viewrope,chen2025vrag} primarily model the relation between memory features and the current frame being generated with an attention mechanism. They  perform frame-level retrieval based on attention similarity. Their memory representations remain at frame granularity and are insufficient for achieving fine-grained spatio-temporal consistency. Hong et~al.~\cite{hong2025relic} introduce a learnable retrieval mechanism, but it fails to scale video generation to minute-level.

\noindent\textbf{Long Video Extrapolation.}
Constrained by the context length seen during training, long-video generation inevitably faces extrapolation beyond the training horizon at inference. Existing works fall into three main categories. First, some full sequence diffusion approaches apply training-free strategy~\citep{qiu2023freenoise,lu2024freelong,kim2024fifo-diffuion} to decompose long-video synthesis into overlapping clips generation. This addresses inter-clip smoothness but fails to model long-range dependencies across clips. Second, recent autoregressive methods~\citep{chen2024diffusion_forcing,huang2025self_forcing,chen2025skyreels,teng2025magi1,yi2025deepforcing,liu2025rollingforcing,li2026rollingsink,yang2025longlive,cui2026lol,yin2025causvid} adopt sliding-window inference to limit the computational cost. However, this bounded window attention still discards substantial fine-grained history. Third, another line of work directly extends the context of pretrained full-sequence diffusion models~\citep{wan2025wan,kong2024hunyuanvideo} to generate a full sequence in a single pass. For instance, RIFLEx~\citep{zhao2025riflex} adjusts the frequency parameters of RoPE to alleviate content repetition. UltraViCo~\citep{zhao2025ultravico} introduces weight decay to improve visual quality. However, they do not consider long-term memory when inference length scales. This leads to a pronounced degradation of global spatio-temporal consistency under long-horizon extrapolation. In contrast, our proposed method focuses on efficient information extraction from the long historical context, while mitigating attention dispersion for quality retention.

%% file: sec/3_method.tex
\section{Method}
\label{sec:method}
In this section, we first present the preliminaries of autoregressive video generation, followed by action-conditioned world modeling (\cref{sub-sec:Pre}). \cref{sub-sec:attn_disperison} analyzes the attention dispersion phenomenon that emerges as world models extrapolate over long horizons. To address this limitation and efficiency problem, we introduce a novel decoupled memory architecture, named DecMem, consisting of a Sparse Global Memory (SGM) for efficient long-context modeling and an Anchored Local Memory (ALM) for stable attention distribution. \cref{sub-sec:SGM} and \cref{sub-sec:ALM} provide the details of SGM and ALM respectively. Finally in~\cref{sub-sec:PE}, we present a multimodal position embedding for encoding camera pose with spatio-temporal information.

\subsection{Preliminaries}
\label{sub-sec:Pre}
\noindent\textbf{Autoregressive Video Generation.} Modern video generative frameworks typically operate in a compressed latent space. A pretrained Variational Autoencoder (VAE) encodes the raw video sequence into a latent representation $\mathbf{z}_0^{1:T} \in \mathbb{R}^{C \times T \times H \times W}$. The objective of an autoregressive video generative model is to predict the subsequent latent $\mathbf{z}_0^{T+1}$ conditioned on the denoised history $\mathbf{z}_0^{1:T}$. During the training phase, following Rectified Flow~\citep{liu2022rectiflow}, we sample noise $\epsilon^{T+1} \sim \mathcal{N}(\mathbf{0}, \mathbf{I})$ and construct the noisy latent $\mathbf{z}_t^{T+1}$ via linear interpolation between the clean latent and the noise. We apply teacher forcing~\citep{kondratyuk2023videopoet} paradigm and provide clean history $\mathbf{z}_{0}^{1:T}$ during training. The model $\mathbf{v}_\theta$ is optimized to predict the flow velocity $\mathbf{v}^{T+1} = \epsilon^{T+1} - \mathbf{z}_0^{T+1}$ by minimizing the following objective:
\begin{equation}
    \mathcal{L} = \left\| \mathbf{v}_{\theta}\left(\mathbf{z}_{0}^{1:T}, \mathbf{z}_t^{T+1}, t \right) - \mathbf{v}^{T+1} \right\|_2^2
\end{equation}

\noindent\textbf{Action-Conditioned World Modeling.} To transform a video generator into a world model, we incorporate action condition as control signals. Following the spirit of Hunyuan-Gamecraft~\citep{li2025hunyuan-gamecraft}, the action embedding $\mathbf{a}$ is mapped with a light-weight fusion module $\psi(\cdot)$ and added with video latents:
\begin{equation}
\mathbf{x} = \text{Patchify}(\mathbf{z}) \oplus \psi(\mathbf{a})
\end{equation}
where $\oplus$ denotes the element-wise addition. The feature $\mathbf{x}$ is then sent to Transformer blocks for further fusion. This ensures deep multimodal fusion between visual features and action controls, while keeping negligible computational overhead.

\subsection{Attention Dispersion in Long World Simulation}
\label{sub-sec:attn_disperison}
\begin{figure}[h] 
    \centering
    \begin{minipage}[t]{0.65\linewidth}
        \centering
        \includegraphics[width=\linewidth]{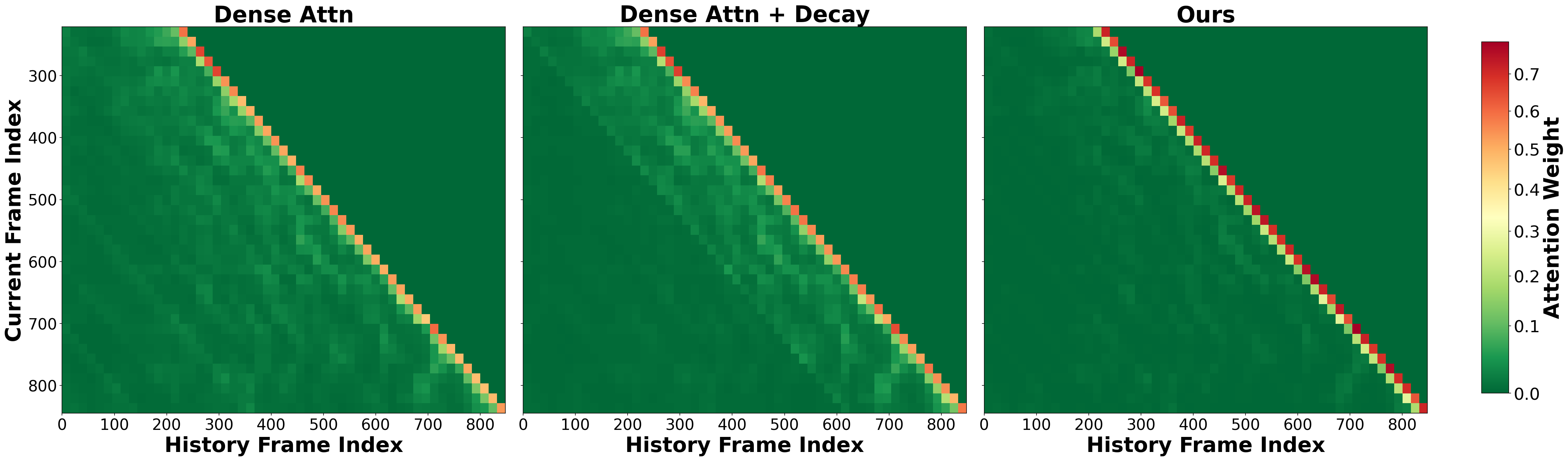}
        \caption{Attention maps of different long world modeling approaches during long-horizon video inference.}
        \label{fig:attn_map}
    \end{minipage}
    \hfill 
    \begin{minipage}[t]{0.3\linewidth}
        \centering
        \includegraphics[width=\linewidth]{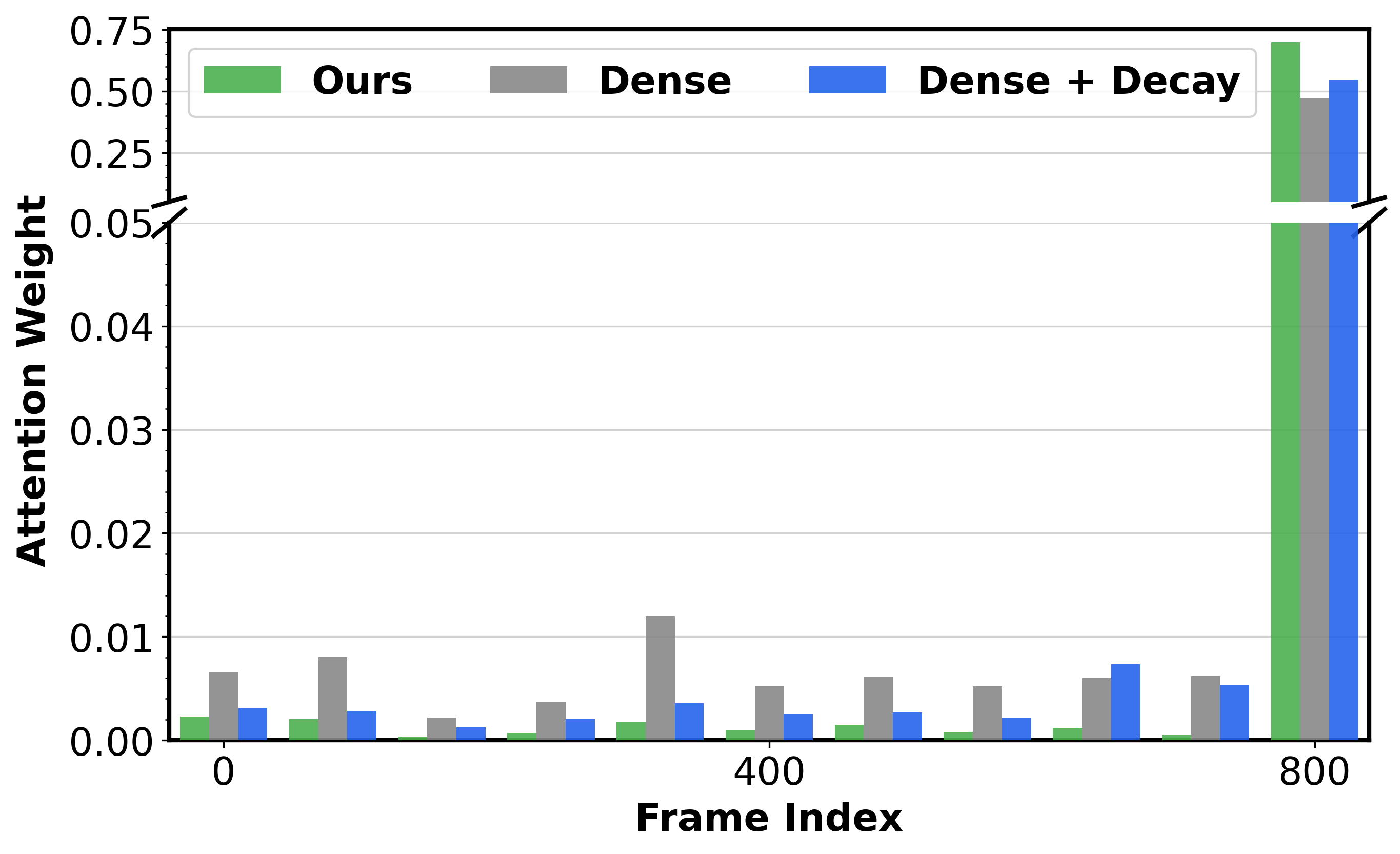}
        \caption{Attention distribution in the generation of the $810{th}$ frame (sampled every 80 frames).}
        \label{fig:attn_dis}
    \end{minipage}
\end{figure}

In this section, we first analyze the attention mechanism and identify the root cause of failure in naïve long-video inference. This analysis naturally motivates our new architectural design. As shown in~\cref{fig:intro}(a), the naïve dense attention architecture exhibits pronounced quality degradation in long extrapolation. Conversely, training-free decay strategies~\citep{zhao2025ultravico} suppress out-of-window attention to mitigate short-range distortion, but doing so at the expense of long-term spatio-temporal consistency.

To analyze the attention mechanism, we visualize the attention maps throughout a long-horizon inference in~\cref{fig:attn_map}. As the extrapolation length grows, it can be observed that the query's attention is progressively diluted by a pool of historical tokens. A vast number of historical features acquire small but non-zero weights. This phenomenon becomes particularly pronounced in the generation of the $810{th}$ frame (see~\cref{fig:attn_dis}). The resulting long-tail distribution inevitably lowers the effect weights allocated to those semantically critical historical frames (see~\cref{sec:more_attn_disperison} for more details).

Mechanically down-weighting distant tokens~\citep{zhao2025ultravico} indiscriminately suppresses all out-of-window attention (\cref{fig:attn_map}) to emphasize the within-window attention, eliminating genuine long-range dependencies, and thereby cutting off the model's access to distant critical information. This dilemma surfaces a key insight: what is required is not a more carefully engineered attention prior, but \emph{a learnable architecture that adaptively suppresses redundancy to preserve attention concentration, while explicitly extracting and exploiting history features that helps long-term memory retention.}

Building on this analysis, we propose a novel \textbf{dec}oupled \textbf{mem}ory architecture for efficient long-range memory and resistance to attention dispersion. To this end, we design a Sparse Global Memory (SGM) module (\cref{sub-sec:SGM}) for efficient and fine-grained memory access and an Anchored Local Memory (ALM) module (\cref{sub-sec:ALM}) to relieve the attention dispersion problem. The outputs of these two modules are fused through a learnable gating mechanism, adaptively preserving short-range fidelity and long-range memory.

\label{method}
\begin{figure}
    \centering
    \includegraphics[width=0.98\linewidth]{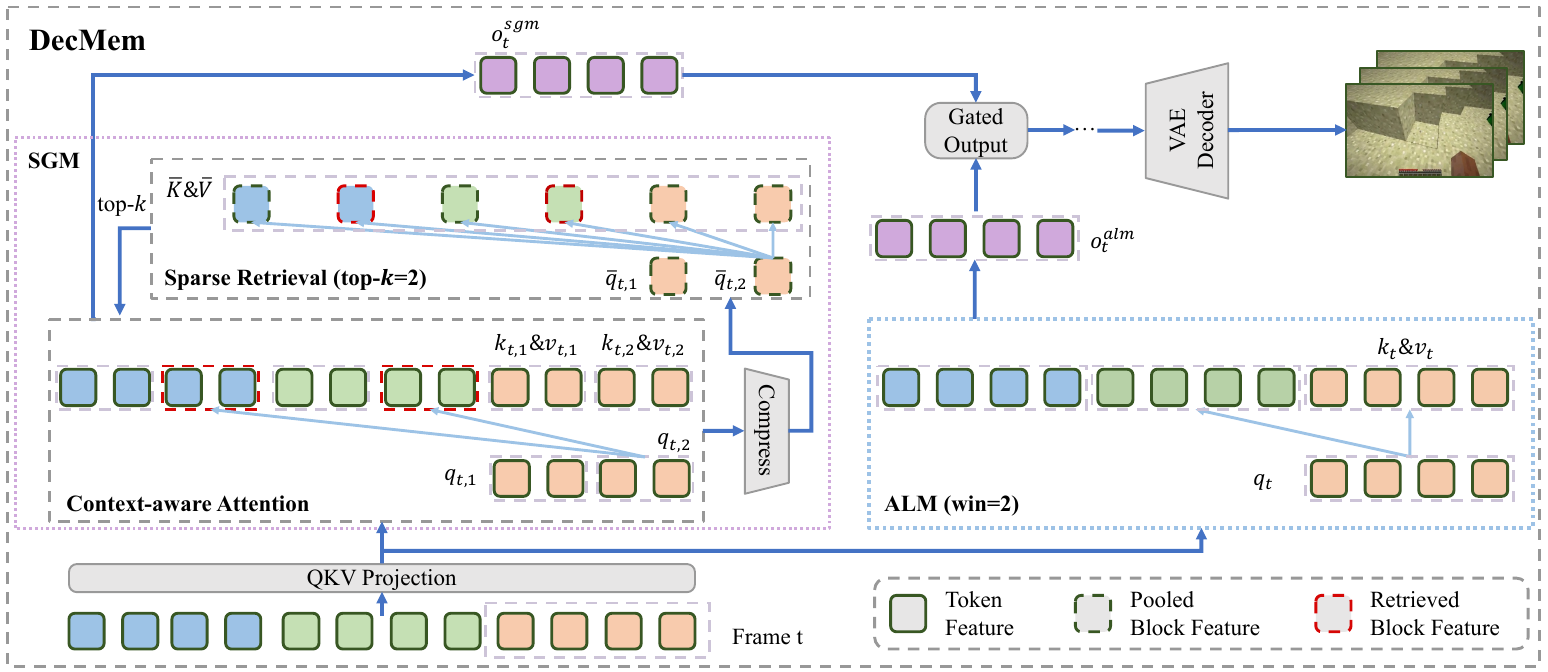}
    \caption{DecMem pipeline comprises decoupled memory for long-term consistency and extrapolation generalization while keeping the computational cost low. Sparse Global Memory (SGM) combines a block-level sparse retrieval module and a context-aware attention module for long-term memory fine-grained retrieval in an end-to-end manner, whereas Anchored Local Memory (ALM) keeps short-term transition smooth. For clearer visualization, we display 3 frame latents as key \& value and the last frame indexed by $t$ as query. Each frame contains 2 blocks with 2 tokens per block.}
    \label{fig:pipe}
\end{figure}

\subsection{Sparse Global Memory}
\label{sub-sec:SGM}
To overcome the scaling bottleneck of dense attention in long-video synthesis, we introduce the Sparse Global Memory (SGM) module. By executing retrieval at a fine-grained block level, SGM enables precise recall of long-term dependencies without the heavy computational overhead of global modeling. 

Specifically, SGM carries out a two-stage process, namely block-level sparse retrieval and context-aware attention (see \cref{fig:pipe}). In block-level sparse retrieval, we first split the latent frame into $M$ non-overlapping blocks and aggregate features within each block by pooling. These pooled features are then used to identify the most relevant historical blocks to represent the fine-grained memory context. Let $\bar{q}_{t,i}$ denote the pooled feature of the $i$-th block $q_{t,i}$ in the current frame $t$. We evaluate the relevance between this block and the historical blocks by computing their attention scores using the pooled features. Blocks with the top-$k$ scores are chosen to represent the fine-grained memory context $\mathcal{C}_{t, i}$ for $q_{t,i}$. 

After block-level sparse retrieval, we next perform context-aware attention using the retrieved blocks in $\mathcal{C}_{t, i}$. This helps to reduce attention computation from the full sequence to only the top-$k$ most relevant blocks, preventing the per-step computation from growing linearly. Specifically, we perform a dense attention computation for each token in the query block $q_{t,i}$ with every token in the retrieved blocks in $\mathcal{C}_{t, i}$.
Once this block-level attention computation is completed for every query block, we assemble the block outputs into a frame output $o^{\rm sgm}_t$ for the current frame $t$.

Through SGM's sparse block-level computations, we substantially reduce the attention cost while achieving fine-grained retrieval over long-range global history.


\subsection{Anchored Local Memory}
\label{sub-sec:ALM}
To counteract quality degradation caused by attention dispersion during extended inference, we introduce Anchored Local Memory (ALM) as a complementary branch to stabilize the attention distribution. Given that temporally adjacent frames inherently exhibit the strongest visual and semantic correlation with the current frame, ALM strictly confines its attention to a local window of the most recent frames, thereby providing a high-confidence attention anchor to mitigate temporal drift and reinforce the model's long-range extrapolation capability.

Specifically, we implement ALM with a sliding window attention mechanism, with the context window constrained to the immediate past $w$ frames (see \cref{fig:pipe}). %
This formulation explicitly models the interaction between the current frame and its immediate history, stabilizing the attention distribution during variable-length extrapolation.

Finally, we adaptively fuse the outputs of the two branches through a learnable gating mechanism, with the ALM output $o^{\rm alm}_t$ serving as a stable baseline that prevents the fused attention from being diluted by long-tail distractors from distant frames, and the SGM output $o^{\rm sgm}_t$ endowing the model with fine-grained memory and retrieval capability over previously visited scenes:
\begin{equation}
    o_t = o^{\rm alm}_t + G_t \odot o^{\rm sgm}_t
\end{equation}
where $G_t$ is the learnable gate derived from the current frame features. Modulated by the gating, the two branches jointly yield an adaptive trade-off between global consistency and extrapolation robustness, sustaining spatio-temporal consistency in minute-long video generation without collapse.

\subsection{Multimodal Position Embedding}
\label{sub-sec:PE}
To inject geometric and spatio-temporal priors into the attention computation, we extend video RoPE~\citep{wan2025wan, kong2024hunyuanvideo,yang2024cogvideox} by incorporating camera geometry embeddings. To avoid modality interference in the feature subspace, we partition the channels and apply position embeddings separately to each group. We follow PRoPE~\citep{li2025cameras_prope} to inject the camera geometry $P$ for encoding the relative geometric relationship. For a token at position $(t_i, x_i, y_i)$ (i.e., the $i$-th token, located in frame $t_i$ at patch coordinates $(x_i, y_i)$), the full transformation matrix $\mathbf{R}_{full}^{(i)}$ can be written as:
\begin{equation}
    \label{eq:rotation_matrix}
    \mathbf{R}_{full}^{(i)} = \text{diag}(\mathbf{R}_{cam}, \mathbf{R}_{sp}, \mathbf{R}_{tem}) = 
    \begin{bmatrix} 
    \mathbf{R}_{cam}(P_{t_i}) & \mathbf{0} & \mathbf{0} \\
    \mathbf{0} & \mathbf{R}_{sp}(x_i,y_i) & \mathbf{0} \\
    \mathbf{0} & \mathbf{0} & \mathbf{R}_{tem}(t_i)
    \end{bmatrix}
\end{equation}
where $\mathbf{R}_{cam}$, $\mathbf{R}_{sp}$, and $\mathbf{R}_{tem}$ denote the transformation matrices derived from camera parameters, patch coordinates, and frame index respectively. Then the position-encoded query and key can be computed as:
\begin{equation}
\label{eq:pe_equation}
q_i = (\mathbf{R}_{full}^{(i)})^{\!\top}\,\mathrm{Proj}_q(h_i), \quad 
k_i = (\mathbf{R}_{full}^{(i)})^{-1}\,\mathrm{Proj}_k(h_i).
\end{equation}
where $h_i$ denotes $i$-th token of input feature in each spatio-temporal attention layer, $\mathrm{Proj}_q(.)$ and $\mathrm{Proj}_k(.)$ are the respective projection transformations of the input features. More details can be found in~\cref{prope}. This multimodal position embedding jointly encodes camera geometry, patch location and frame index, bringing precise geometric and spatio-temporal perception.


%% file: sec/4_experiment.tex
\section{Experiment}
\label{sec:exp}

\textbf{Implementation details.}
\label{sec:imple_details}
Our pipeline is implemented on a 1B pretrained video generation model in chunk-wise auto-regressive manner, with each chunk containing 4 latents for faster generation. The latents within a chunk can attend to each other, and we keep the causality between chunks. We train our model on 64 NVIDIA H200 GPUs with a global batch size of 64. For long-term memory retrieval in our SGM module, we divide each frame into 6 blocks of the same size with padding. We set $k$ in top-$k$ historical blocks retrieval to 80 unless otherwise specified. Our ALM module employs a context window of 8 frame latents. We train DecMem on the WorldMem~\citep{xiao2025worldmem} datasets and apply FID~\citep{heusel2017rfid}, PSNR, and LPIPS~\citep{zhang2018lpips} to evaluate the distribution-level, pixel-level, and perceptual-level similarity between generated results and the ground truth. More details can be found in~\cref{sec:more_details}.

\noindent\textbf{Baselines.}
We compare our DecMem with Oasis~\citep{oasis2024}, MineWorld~\citep{guo2025mineworld}, and WorldMem~\citep{xiao2025worldmem} to demonstrate the effectiveness of our method. These methods are trained fully on the MineCraft datasets and hence have abundant domain knowledge. Oasis and MineWorld both use sliding windows to handle their memory, whereas WorldMem employs FOV-based memory retrieval. 

\subsection{Quantitative Experiments}
\label{sub-sec:quantitative exp}
\noindent\textbf{Evaluation Settings.} We evaluate our method and the baseline models in handling controllable video generation within training context and beyond. All the models are provided with 221 ground-truth frames as memory bank initialization, and tasked to generate the subsequent 120 frames. For Oasis and MineWorld with a context window of $w$ frames (with $w$ being 8 and 32 respectively), we initialize their memory frames with the $w - 1$ most recent frames. For WorldMem, which keeps 8 frames in its sliding window, we additionally keep other previous frames in its memory bank following its original setting. For our method with end-to-end memory retrieval, all the frames are fed into the model for fine-grained block retrieval. More details can be found in~\cref{sub:more_train_eval_details}.

\noindent\textbf{Within Training Window.}
Here, we use the first 8 generated frames (i.e., $222{nd}$--$229{th}$ frames) to assess the proficiency of each model in retrieving and leveraging immediate historical context. This comparison ensures that the inference remains within the respective training window of each model. As shown in~\cref{tab:sota}, our method outperforms all other baselines under all the metrics being considered, demonstrating the effectiveness of our precise memory in short term.

\noindent\textbf{Extrapolation Generalization.} 
Here, we use the last 8 generated frames (i.e., $334{th}$--$341{st}$ frames) to evaluate the extrapolation capability of the world models beyond the training length. As illustrated in~\cref{tab:sota}, our method demonstrates superior robustness during extrapolation, effectively preserving spatial consistency. In contrast, competing baselines such as WorldMem exhibit rapid performance degradation after crossing the training-length threshold. 

\noindent\textbf{User Study.} 
To validate the effectiveness of our method from a perceptual perspective, we conducted a user study with 58 participants, who were asked to rate the generated videos along three dimensions: Visual Quality (VQ), Action Controllability (AC), and Spatio-temporal Consistency (STC). As reported in~\cref{tab:sota}, our method outperforms all baselines across three dimensions, validating DecMem's overall superiority in visual quality, controllability, and spatio-temporal consistency.

\noindent\textbf{Inference Latency.} We also compare the inference speed by computing frame rates (in FPS) from the average generation time for 120 frames. The results in~\cref{tab:sota} show that our method outperforms all the baselines in efficiency, achieving nearly 2x speedup compared with the most competitive baseline.

\begin{table}[htbp]
\centering
\caption{Quantitative comparison and user study results.}
\resizebox{0.9\linewidth}{!}{
\begin{tabular}{l ccc ccc ccc c}
\toprule
& \multicolumn{3}{c}{\textbf{Within Training Window}} & \multicolumn{3}{c}{\textbf{Extrapolation Generalization}} & \multicolumn{3}{c}{\textbf{User Study}} &  \\
\cmidrule(lr){2-4} \cmidrule(lr){5-7}  \cmidrule(lr){8-10} 
\textbf{Method}
& PSNR$\uparrow$ & LPIPS$\downarrow$  &FID$\downarrow$ 
& PSNR$\uparrow$ & LPIPS$\downarrow$  &FID$\downarrow$ & VQ$\uparrow$ & AC$\uparrow$ &  STC$\uparrow$ & \multirow{-2}{*}{FPS$\uparrow$} \\
\midrule
MineWorld~\citep{guo2025mineworld} & 20.2989 & 0.1789 & 41.9661 & 14.6109 & 0.3736 & 74.2075  & 15.06\% & 14.49\% & 14.29\% & 0.1588 \\
Oasis~\citep{oasis2024} & 24.1293 & 0.1196 & 15.9163 & 13.4232 & 0.4475 & 63.8851 & 25.87\% & 22.26\% & 19.41\% & 1.9806 \\
WorldMem~\citep{xiao2025worldmem} & 26.5414 & 0.0797 & 11.7379 & 19.1401 & 0.2673 & 38.4689 & 19.31\% &  25.33\% & 24.16\% & 0.5424 \\
Ours & \textbf{30.0785} & \textbf{0.0494} & \textbf{9.8904} & \textbf{25.2294} & \textbf{0.1006} & \textbf{16.2667} & \textbf{39.77\%} & \textbf{37.81\%} & \textbf{42.12\%}  & \textbf{3.6496} \\
\bottomrule
\end{tabular}}
\label{tab:sota}
\end{table}

\subsection{Qualitative Experiments}
\label{qualitative exp}
\begin{figure}[t]
    \centering
    \includegraphics[width=0.9\linewidth]{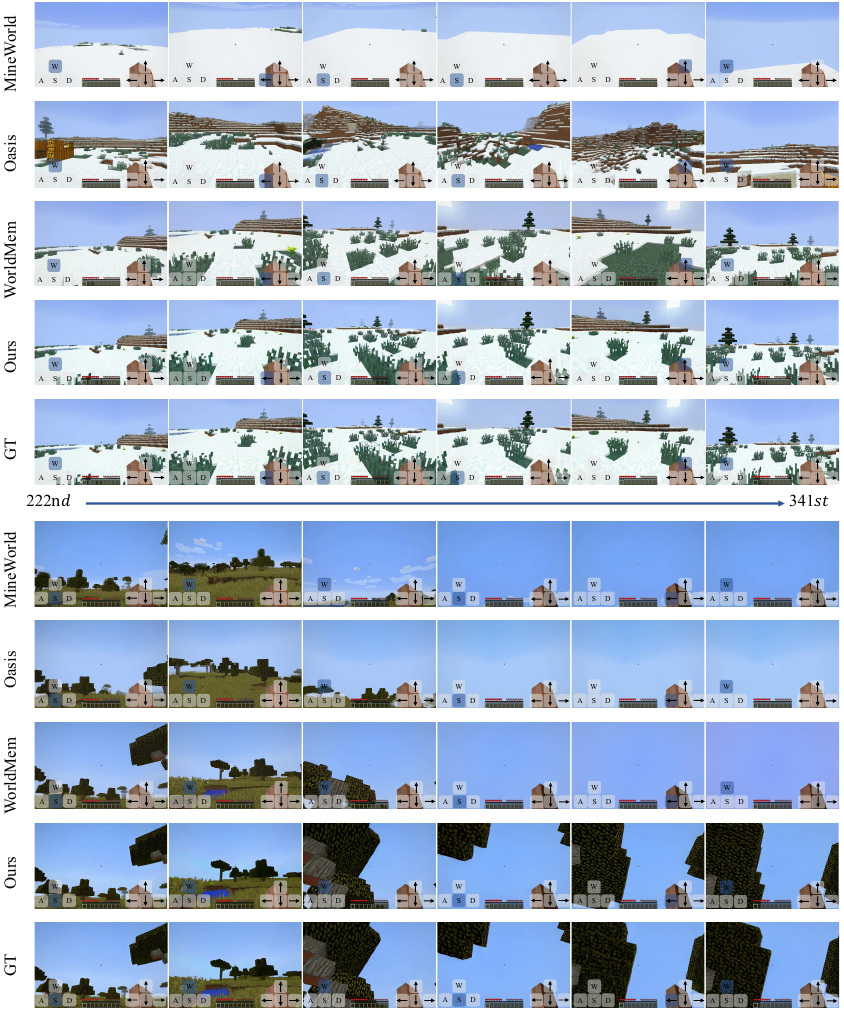}
    \caption{Qualitative comparison on the Minecraft Datasets.}
    \label{fig:sota_com}
\end{figure}

Following the setting of~\cref{sub-sec:quantitative exp}, we initialize the environment with 221 frames and let the models generate the subsequent 120 frames to show their spatio-temporal consistency with the initial environment. ~\cref{fig:sota_com} demonstrates that our method achieves superior fidelity in short-term generation by precisely reconstructing local details, while other methods (including FOV-based WorldMem) struggle to maintain the accuracy of detailed memory. More importantly, in long-term scenarios, our method effectively preserves fine-grained details and overall video quality, ensuring robust spatial-temporal consistency that surpasses existing baselines. 

\begin{figure}[ht]
    \centering
    \includegraphics[width=0.9\linewidth]{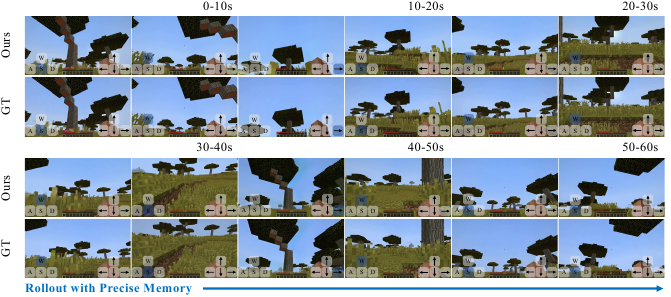}
    \caption{Minute-long video generation results with precise memory.}
    \label{fig:longterm}
\end{figure}

Notably, our method supports ultra-long video synthesis of minute-long duration (see~\cref{fig:longterm}) while maintaining rigorous consistency in revisiting scenes, effectively overcoming temporal degradation common in long-horizon generation.
\vspace{-10pt}



\subsection{Ablation Study}
\begin{figure}[htbp]
    \centering
    \includegraphics[width=0.98\linewidth]{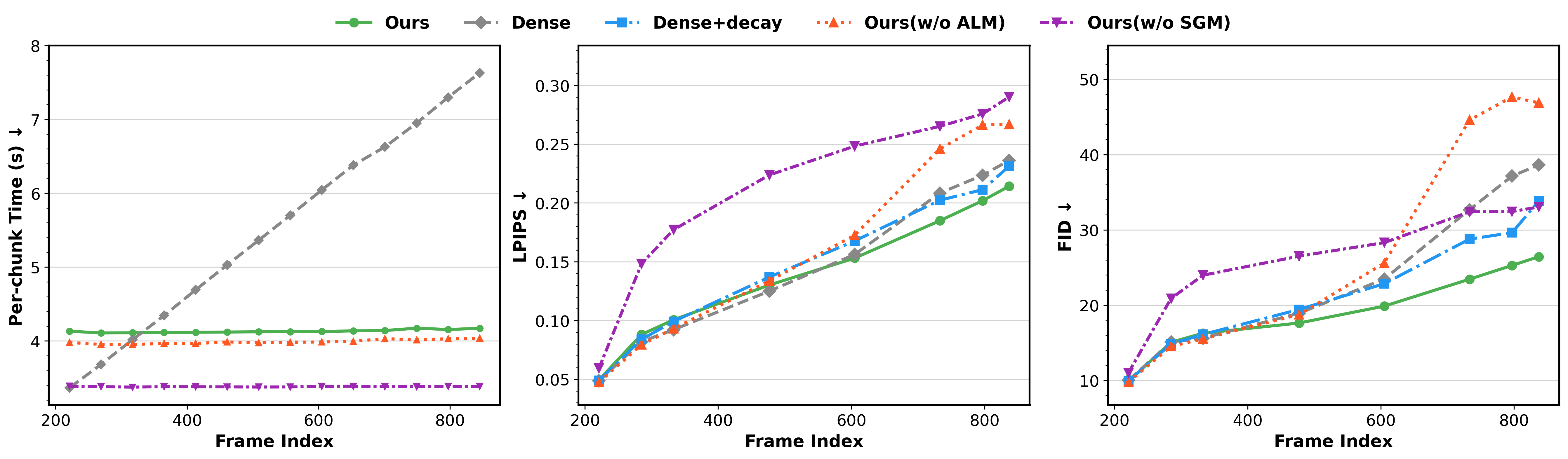}
    \caption{Quantitative comparison of efficiency and quality between different design. (Left) Time to generate one chunk at the current frame index. (Middle, Right) LPIPS and FID computed using 8 neighboring frames (t→t+8) at each position.}
    \label{fig:com_metric}
\end{figure}

\textbf{Component ablation.}
To validate the contribution of each module in DecMem, we conduct component ablation by removing SGM and ALM separately. Each variant is initialized with 221 memory frames and tasked to generate over 600 frames. We additionally compare against Dense Attention and Dense Attention with a training-free temporal decay strategy~\citep{zhao2025ultravico}. The following points are observed from the results reported in \cref{fig:com_metric}. 
\textbf{(1) Dense Attention} exhibits linearly growing latency (\cref{fig:com_metric}, left) and quality collapse in long extrapolation, revealing its fundamental inability to scale to long-horizon generation.  \textbf{(2) Dense Attention + Decay} alleviates late-stage degradation (after 700 frames) 
but introduces a regression in the middle extrapolation range (around 300$th$-700$th$ frames) as reflected in worse LPIPS scores relative to the dense baseline. This shows uniform temporal decay indiscriminately suppresses both redundant and informative historical features, eroding memory fidelity. 
\textbf{(3) w/o SGM} yields the worst generation quality across the entire extrapolation horizon. Without global memory retrieval, the model degenerates into a local-context-only generator and rapidly loses long-range consistency. \textbf{(4) w/o ALM} preserves reasonable quality in the early extrapolation stage but suffers from severe degradation beyond 600 frames, with FID and LPIPS both worst than that of vanilla Dense Attention. This confirms that, without the local anchoring mechanism, attention dispersion over the growing global context becomes the dominant failure mode, corroborating our analysis in~\cref{sec:intro}. \textbf{(5) Full DecMem} matches Dense Attention in the early stage and consistently surpasses all variants in the later stage, while maintaining a near-constant computational cost (thanks to the sparse block retrieval). Only the full model, which combines sparse global retrieval with local anchoring, maintains stable quality throughout the entire rollout.

\noindent\textbf{Number of retrieval blocks (top-$k$).}
In this section, we compare different numbers of memory retrieval blocks. We initialize the model with 221 memory frames and evaluate it under three settings: (a) within training window ($222{nd}$--$229{th}$ frames), (b) mid-range extrapolation ($334{th}$--$341{st}$ frames), and (c) long-range extrapolation ($798{th}$--$805{th}$ frames), with $k$ set to 20, 50, 80, and 100. As shown in~\cref{tab:topk_ablation}, increasing $k$ does not yield consistent improvements across all metrics. Notably, increasing it from 80 to 100 degrades both PSNR and FID under long-range extrapolation. Since $k$ governs both the recall coverage of SGM and the consistency quality after fusion with the short-range anchored signals from ALM, an overly large value dilutes the retrieved context and weakens this complementarity. We therefore set $k$ to 80 to balance long- and short-range quality.

\begin{table}[htbp]
\centering
\caption{Ablation on Number of retrieval blocks}
\scalebox{0.9}{
\begin{tabular}{l ccc ccc ccc}
\toprule
& \multicolumn{3}{c}{\textbf{Within Training Context}} & \multicolumn{3}{c}{\textbf{Mid-range Extrapolation}} & \multicolumn{3}{c}{\textbf{Long-range Extrapolation}} \\
\cmidrule(lr){2-4} \cmidrule(lr){5-7} \cmidrule(lr){8-10}
$k$ & PSNR$\uparrow$ & LPIPS$\downarrow$  & FID$\downarrow$ & PSNR$\uparrow$ & LPIPS$\downarrow$ & FID$\downarrow$ & PSNR$\uparrow$ & LPIPS$\downarrow$  & FID$\downarrow$ \\ 
\midrule
20 & 29.6230 & 0.0522 & 10.2337 & 24.3749 & 0.1105 & 16.9071 & 19.6497 & 0.2217 & 27.7425\\
50 & 29.9877 & 0.0500 & 9.9851 & 25.0605 & 0.1023  & 16.2642 & 19.9425 & 0.2100 & 27.0902  \\
80 & 30.0785 & 0.0494 & \textbf{9.8904} & 25.2294 & 0.1006 & 16.2667 & \textbf{20.5896} & 0.2019 & \textbf{25.2748} \\
100 & \textbf{30.1529} & \textbf{0.0490} & 9.9653 & \textbf{25.4616} & \textbf{0.0962} & \textbf{15.5984} & 20.5535 & \textbf{0.1994} &  25.8790 \\
\bottomrule
\end{tabular}}
\label{tab:topk_ablation}
\end{table}

%% file: sec/5_conclusion.tex
\section{Conclusion}
This paper proposes a fine-grained, learnable and scalable memory architecture for world models. We first analyze two intrinsic limitations of the na\"ive dense attention design under long-horizon inference, namely computational inefficiency and attention dispersion. Building upon a systematic analysis of the attention dispersion, we propose a decoupled memory architecture, consisting of a Sparse Global Memory (SGM) branch which performs fine-grained, learnable sparse memory retrieval for efficient long-range memory preservation, and an Anchored Local Memory (ALM) branch which supplies stable attention anchors that effectively counteract dispersion from distant noise. Extensive experiments validate the effectiveness of this architecture, ultimately enabling minute-long, efficient, and highly consistent controllable video generation.

%% file: sec/6_appendix.tex
\section{Implementation Details}
\label{sec:more_details}

\subsection{Experiment Settings.}
\label{sub:more_train_eval_details}
\textbf{Training and Evaluation Details.} 
We train our DecMem on WorldMem~\citep{xiao2025worldmem} datasets, which contains 11 k videos with 1500 frames at 360x640 resolution and 10 FPS. We randomly sample 237 frames and resize them to 352x640 for training and evaluation. We adopt a two-stage training strategy. In the first stage, we initialize from a pre-trained full-sequence video generation checkpoint and adapt its architecture into a causal generation paradigm, training for 25K steps so that the model robustly establishes autoregressive generation as a reliable backbone for subsequent memory injection. In the second stage, we integrate the proposed Sparse Global Memory (SGM) and Anchored Local Memory (ALM) modules on top of this causal backbone, and jointly train for an additional 25K steps so that the model learns to retrieve sparse global context and exploit anchored local memory in a coordinated manner, ultimately delivering fine-grained long-horizon spatiotemporal consistency. We apply the AdamW optimizer with a learning rate of 2e-5 and adopts teacher forcing strategy. The training process lasts for approximately 7 days. For evaluation, we apply 300 videos from the WorldMem~\citep{xiao2025worldmem} datasets ensuring no overlap with the training data. For diffusion-based methods, we keep all of them to denoise 20 steps for fair comparison. 

\begin{figure}[ht]
    \centering
    \includegraphics[width=\linewidth]{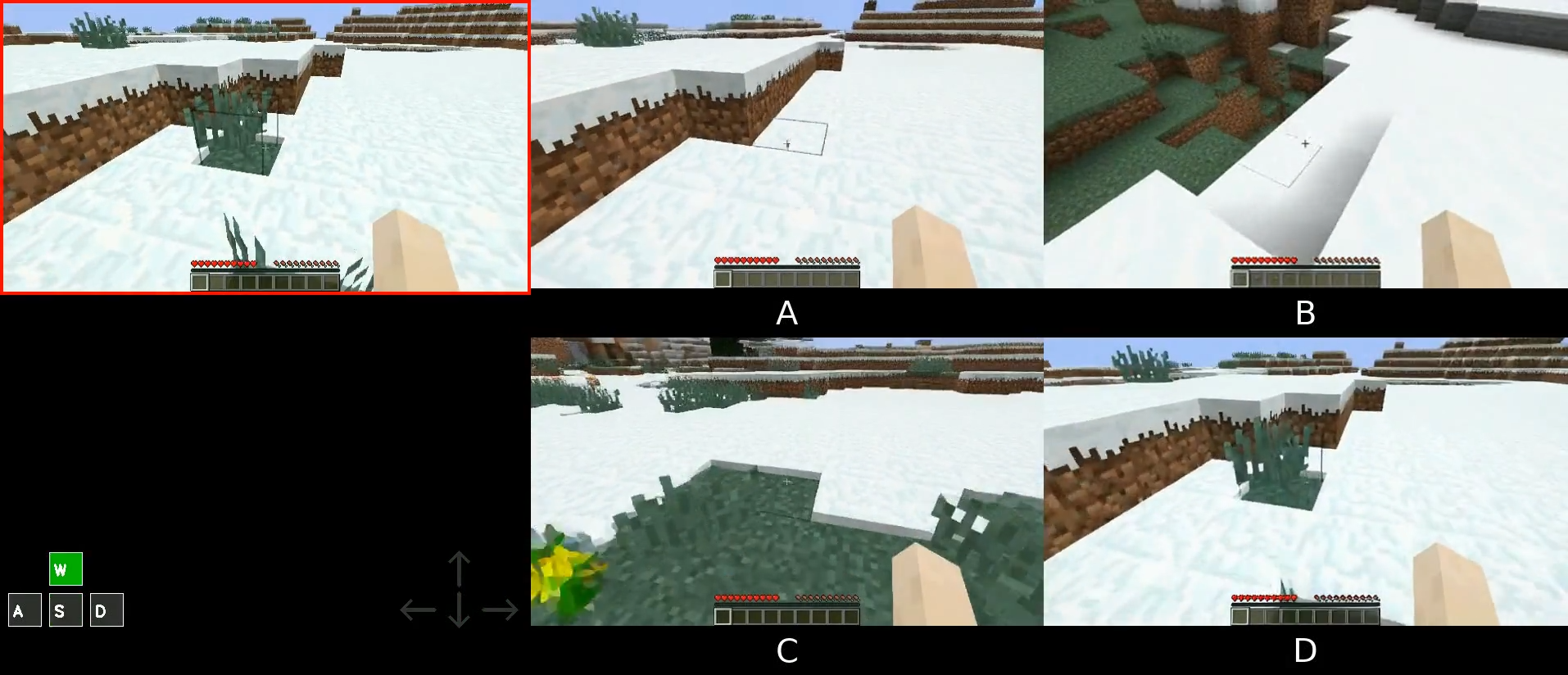}
    \caption{User study demo.}
    \label{fig:user study demo}
\end{figure}

\textbf{User Study Details.}
To assess generation quality from a perceptual standpoint, we conducted a user study with 58 participants. Each trial was presented in a unified layout (see~\cref{fig:user study demo}): the left panel simultaneously showed the ground-truth reference clip together with its corresponding action control signals, while the right panel played, side by side, the candidate videos generated by different methods, which were randomly shuffled and anonymized (labeled A, B, C, and D, respectively) to eliminate positional bias and method-identification cues. Participants were asked to select the indices of the videos they deemed best under each predefined evaluation criterion, i.e, visual quality, action controllability, and spatio-temporal consistency. We aggregated all responses and computed, for every metric, the preference rate of each method; the final results are summarized in~\cref{tab:sota}.

\subsection{Base Model Architecture}
\label{base model}
\begin{figure}[ht]
    \centering
    \includegraphics[width=\linewidth]{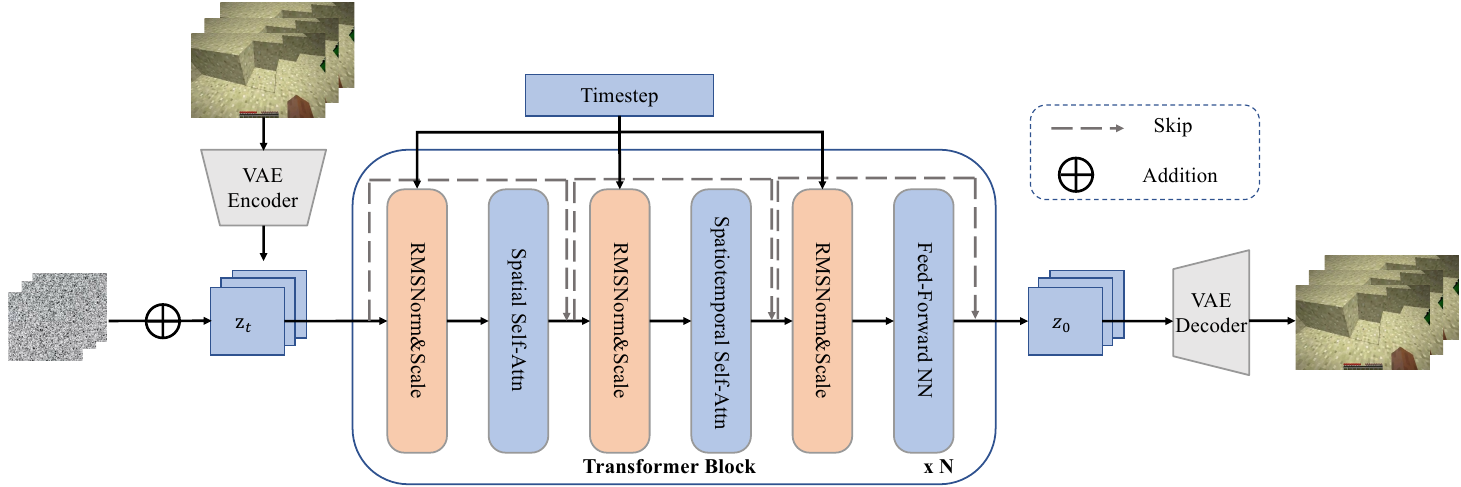}
    \caption{Base Model Architecture.}
    \label{fig:base model}
\end{figure}

For our pretrained video generation models, we apply the latent diffusion transformer as our base model as illustrated in~\cref{fig:base model}. Since we rely on actions and poses to control scene generation rather than using prompts for guidance, we discard the cross-attention module designed for the T2V task, employ spatial self-attention to fuse information within frames, and use spatiotemporal self-attention to capture the relationships among latents across frames. Before each attention or feed-forward network (FFN) module, the timestep is mapped to a scale, which is then used to apply RMSNorm~\citep{zhang2019rmsnorm} to the features.

\subsection{Details of Multimodal Position Embedding}
\label{prope}
In~\cref{sub-sec:PE}, we introduce a multimodal position embedding that injects camera geometry, patch coordinates, and frame indices into the attention computation. Concretely, the per-head dimension of 72 is evenly partitioned into three groups of 24 channels, with each group encoding one modality through its corresponding transformation. 

\textbf{Camera Embedding.}
For the camera-pose channel, we follow PRoPE~\citep{li2025cameras_prope} for projective positional encoding. Let $K_t \in \mathbb{R}^{3\times 3}$ denote the camera intrinsics of the $t$-th frame and $T^{cw}_t = (R^{cw}_t,\, t^{cw}_t) \in \mathrm{SE}(3)$ denote its world-to-camera extrinsics. The standard $3\!\times\!4$ projection matrix that maps a 3D world point to the image plane of camera $t$ is:
\begin{equation}
P_t \;=\; \big[\,K_t \;\; \mathbf{0}_{3\times 1}\,\big]\, T^{cw}_t .
\end{equation}
To make $P_t$ invertible, the standard basis vector $e_4=(0,0,0,1)^\top$ is appended to $P_t$ as its last row, yielding a $4\!\times\!4$ matrix:
\begin{equation}
\tilde{P}_t \;=\;
\begin{bmatrix}
P_t \\[2pt] e_4^\top
\end{bmatrix} \in \mathbb{R}^{4\times 4}.
\end{equation}
The obtained $\tilde{P}_t$ captures the full viewing frustum and hence it can be applied for encoding the complete geometric relationship between camera views. This can be computed as follows:
\begin{equation}
\tilde{P}_{t_1}\,\tilde{P}_{t_2}^{-1}
\;=\;
\begin{bmatrix} K_{t_1} & \mathbf{0} \\ \mathbf{0} & 1 \end{bmatrix}
\,T^{cw}_{t_1}\!\big(T^{cw}_{t_2}\big)^{-1}\,
\begin{bmatrix} K_{t_2}^{-1} & \mathbf{0} \\ \mathbf{0} & 1 \end{bmatrix},
\label{eq:prope-rel}
\end{equation}
which simultaneously models pose and intrinsics differences between two
views. We apply $\tilde{P}_t$ as a block-diagonal transformation on the camera-pose channels:
\begin{equation}
\mathbf{R}_{cam}(P_t) \;=\; \mathbf{I}_{d_{cam}/4} \,\otimes\, \tilde{P}_t
\;\in\; \mathbb{R}^{d_{cam}\times d_{cam}},
\end{equation}
where $d_{cam}$ is the number of feature channels assigned to the camera
modality and $\otimes$ denotes the Kronecker product. Together with~\cref{eq:pe_equation} in the main text, the resulting query--key inner product is
modulated by $\tilde{P}_{t_1}\tilde{P}_{t_2}^{-1}$ as in
Eq.~\eqref{eq:prope-rel}, so attention is conditioned on the relative
camera frustum geometry.

\textbf{Spatial Embedding.}
For the spatial channels with dimension $d_{sp}=24$, we apply the standard 2D
axial RoPE~\citep{su2024rope} on patch coordinates $(x,y)$. The channels are split evenly into two
halves encoding the horizontal and vertical axes respectively:
\begin{equation}
\mathbf{R}_{sp}(x,y) \;=\;
\mathrm{diag}\!\left(\mathbf{R}_{1\mathrm{d}}\!\left(x;\,\tfrac{d_{sp}}{2}\right),\;
\mathbf{R}_{1\mathrm{d}}\!\left(y;\,\tfrac{d_{sp}}{2}\right)\right),
\end{equation}
where $\mathbf{R}_{1\mathrm{d}}(p;d)$ denotes the canonical 1D rotary matrix of
dimension $d$ at position $p$, built from the frequency basis
$\theta_i=\theta_{\text{base}}^{-2i/d}$, $i=0,\dots,d/2-1$.
The resulting query--key inner product depends only on the relative offset
$(x_1\!-\!x_2,\, y_1\!-\!y_2)$, yielding translation-equivariant intra-frame
spatial perception. 

\textbf{Temporal Embedding.}
For the temporal channels with dimension $d_{tem}=24$, we apply 1D RoPE along
the frame index $t$:
\begin{equation}
\mathbf{R}_{tem}(t) \;=\; \mathbf{R}_{1\mathrm{d}}(t;\, d_{tem}),
\end{equation}
which modulates attention by the relative frame distance $t_1\!-\!t_2$.

Apart from modulating the inner product between query and key with position embedding (\cref{eq:pe_equation}), we follow previous work~\citep{miyato2023gta_rope} to inject a relative transformation to the value and the final output for more aligned feature aggregation. However, we do \emph{not} apply such a transformation on the temporal channels to values or outputs. This process is denoted as:
\begin{equation}
    \label{eq:rotation_matrix_appendix}
    \mathbf{R}_{cs}^{(i)} = \text{diag}(\mathbf{R}_{cam}, \mathbf{R}_{sp}, \mathbf{I}) = 
    \begin{bmatrix} 
    \mathbf{R}_{cam}(P_{t_i}) & \mathbf{0} & \mathbf{0} \\
    \mathbf{0} & \mathbf{R}_{sp}(x_i,y_i) & \mathbf{0} \\
    \mathbf{0} & \mathbf{0} & \mathbf{I}
    \end{bmatrix}
\end{equation}

\begin{equation}
    \begin{aligned}
    v_i = (\mathbf{R}^{(i)}_{cs})^{-1} Proj_v(h_i),
    \quad o^{\prime}_i = \mathbf{R}^{(i)}_{cs} o_i
    \end{aligned}
\end{equation}
where $\mathbf{I}$ represents the identity matrix, $Proj_v$ is the value projection transformations of the hidden states, $o_i^{\prime}$ is the position-encoded output features of $i$-th token. 

By explicitly modeling spatiotemporal and geometric relationships, this multimodal RoPE design strengthens the model's spatiotemporal awareness and establishes a reliable prior that underpins the fine-grained memory.


\section{More Analysis about Attention Dispersion}
\label{sec:more_attn_disperison}
\begin{figure}[ht]
    \centering
    \includegraphics[width=\linewidth]{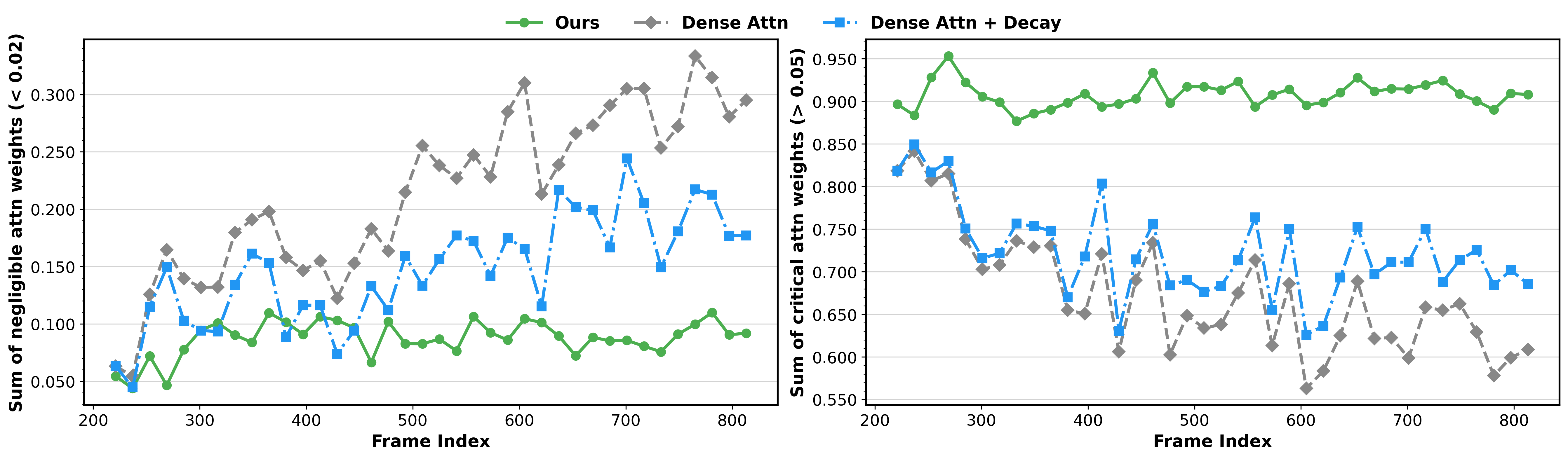}
    \caption{(Left) Sum of negligible attention weights (<0.02) against inference frame index. (Right) Sum of critical attention weights (>0.05) against inference frame index.}
    \label{fig:attn_weight_sum}
\end{figure}

In~\cref{sub-sec:attn_disperison}, we analyze the issue of attention dispersion in long-horizon world modeling. In this section, we further provide a quantitative analysis by examining how the proportions of critical weights and negligible attention weights evolve during inference. As illustrated in~\cref{fig:attn_weight_sum}, for dense attention, the proportion of tail weights gradually increases as inference progresses, while the proportion of key weights correspondingly decreases. This opposing trend leads to the dilution of critical attention.

Although a training-free decay strategy can mitigate the growth of negligible weights and thus help maintain short-term quality, it still exhibits a similar trend to dense attention. Moreover, as analyzed in~\cref{sub-sec:attn_disperison}, it degrades the long-term memory capability of the world model.

In contrast, our method maintains the proportion of irrelevant weights at a relatively constant level throughout inference, thereby reducing the influence of unimportant tail features and preserving a stable share of critical attention weights. This demonstrates the advantage of our decoupled memory design. By introducing ALM as an attention anchor, the model is encouraged to focus most of its attention on important regions, preventing severe quality degradation caused by attention dispersion. Meanwhile, the SGM architecture effectively leverages global memory to explore and utilize long-term temporal features.

\section{Comparison with Industrial-scale Model}
\label{sec:industrial_com}

To further demonstrate the effectiveness of our method, we compare it against two industrial world models, Matrix-Game 2.0~\citep{he2025matrix-game-2} and WorldPlay~\citep{sun2025worldplay}. Both baselines are trained on multi-domain datasets and thus exhibit stronger cross-scene generalization. Besides, they follow Image-to-Video (I2V) or Text-to-Video (T2V) paradigm and do not support video-clip-based memory banks initialization. To guarantee a fair comparison, we deliberately forgo DecMem's advantage of video-conditioned environment initialization and align our input interface with the single-image protocol of the baselines: specifically, we replicate the VAE latent of a single reference frame along the temporal axis to populate the initial chunk that serves as the model's contextual condition.

Each model is tasked to generate 30-second interactive videos conditioned with a single initial image. Owing to insufficient initialization information, ground-truth videos naturally diverge with even identical action sequences, rendering direct comparisons with ground-truth videos meaningless. We therefore adopt the user-study protocol following~\cref{sub-sec:quantitative exp}, performing perceptual evaluations along three perceptual axes: Visual Quality, Action Controllability, and Spatio-temporal Consistency.

As shown in~\cref{tab:industry_com}, our method achieves visual fidelity and action controllability on par with advanced industrial models, while advancing in long-horizon spatio-temporal consistency (+5.14\%). These results demonstrate the effectiveness of our approach for long-term, consistent, and controllable world generation.

\begin{table}[htbp]
\centering
\caption{Results of user study for comparison with industrial world models.}
\scalebox{0.9}{
\begin{tabular}{l ccc}
\toprule
\textbf{Evaluation Criteria}
& Matrix-Game 2.0~\citep{he2025matrix-game-2} & WorldPlay~\citep{sun2025worldplay} 
& Ours \\
\midrule
Visual Quality$\uparrow$ & 28.74\% & 35.04\% & 36.22\% \\
Action Controllability$\uparrow$  & 33.07\% & 29.96\% & 36.96\% \\
Spatio-temporal Consistency$\uparrow$ & 26.07\% & 34.39\% & 39.53\% \\
\bottomrule
\end{tabular}}
\label{tab:industry_com}
\end{table}

\section{More Ablation study}
\begin{figure}[ht]
    \centering
    \includegraphics[width=\linewidth]{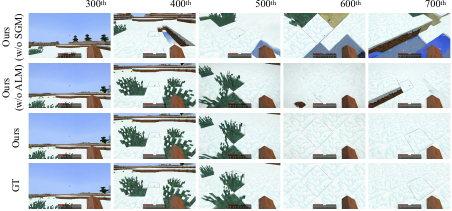}
    \caption{The visualization results to show the effectiveness of each components.}
    \label{fig:ablation_vis}
\end{figure}

\textbf{Visualization of the effectiveness of each Module}
To further validate the efficacy of each component, we compare DecMem against a series of ablated variants and qualitatively analyze the generated samples. Following the protocol in~\cref{sub-sec:quantitative exp}, we initialize the memory bank with 221 frames and let each model auto-regressively roll out the subsequent 500 frames. As shown in~\cref{fig:ablation_vis}, \textbf{(1) w/o SGM}, deprived of the sparse global retrieval mechanism, can only attend to nearby frames, and once generation exceeds the local context window, long-range memory collapses entirely, causing the output to drift markedly away from the ground truth  (GT) and manifest as pronounced scene-identity drift. \textbf{(2) w/o ALM} initially preserves short-range fidelity, however, the generation quality deteriorates sharply beyond roughly 600 frames, with high-frequency details lost and the underlying scene geometry collapsing, revealing that without anchored local memory the model is prone to drift in long-term because of the attention dispersion. In contrast, \textbf{(3) Full DecMem} simultaneously preserves fidelity in short-range and supports stable long-horizon extrapolation, ultimately delivering fine-grained, spatiotemporally consistent minute-long video generation. These observations directly corroborate our central claim that the global consistency and long-range extrapolation fidelity can be addressed through a decoupled memory architecture.

\noindent\textbf{Action classifier free guidance}
Inspired by text-conditioned visual generation models that use classifier-free guidance (CFG) to adjust generation diversity and adherence to the text, we explored the impact of applying CFG to actions in world models on image quality. Specifically, during training, we randomly set the conditional action embeddings to zero and added them to the original latent, simulating the approach of training with dropped actions. 
During inference, the model predicts the flow velocity $\mathbf{v}_{\theta}(\mathbf{z}_t, a_t, t)$ based on action $\mathbf{a}_t$ for each latent $\mathbf{z}_{t}$ at step t. When CFG is applied, we first obtain the conditioned and unconditioned predictions, $\mathbf{v}_{\theta}(\mathbf{z}_t, a_t, t)$ and $\mathbf{v}_{\theta}(\mathbf{z}_t, \varnothing, t)$. The final guided velocity is computed as a weighted combination:
\begin{equation}
    \hat{\mathbf{v}}_{\theta}(\mathbf{z}_t, a_t, t) = \mathbf{v}_{\theta}(\mathbf{z}_t, \varnothing, t) + s\cdot(\mathbf{v}_{\theta}(\mathbf{z}_t, a_t, t)-\mathbf{v}_{\theta}(\mathbf{z}_t, \varnothing, t))
\end{equation}
where $s$ denotes the guidance scale. Notably, for the sake of brevity, we omit other conditional information and illustrate the auto-regressive denoising process for a single frame only.

Following the experimental setup in~\cref{sub-sec:quantitative exp}, we compare the quantitative performance of the two methods. For CFG branch, we apply a guidance scale of 7.5 in the denoising process. As shown in~\cref{tab:cfg}, disabling CFG yields higher pixel fidelity (PSNR) within the training horizon and during the early stage of extrapolation. However, as the extrapolation length grows, the generation quality of the CFG-free model degrades rapidly and eventually suffers from large distribution difference from ground truth, whereas enabling CFG substantially keeps both the stability and the generation quality under long-horizon extrapolation. This observation suggests that CFG trades a marginal loss in short-range fidelity for a pronounced gain in long-range quality.

\begin{table}[htbp]
\centering
\caption{Ablation on action classifier-free guidance.}
\scalebox{0.8}{
\begin{tabular}{l ccc ccc ccc}
\toprule
& \multicolumn{3}{c}{\textbf{Within Training Context}} & \multicolumn{3}{c}{\textbf{Middle Extrapolation}} & \multicolumn{3}{c}{\textbf{Long Extrapolation}} \\
\cmidrule(lr){2-4} \cmidrule(lr){5-7} \cmidrule(lr){8-10}
\textbf{Method} & PSNR$\uparrow$ & LPIPS$\downarrow$  & FID$\downarrow$ & PSNR$\uparrow$ & LPIPS$\downarrow$  & FID$\downarrow$ & PSNR$\uparrow$ & LPIPS$\downarrow$  & FID$\downarrow$ \\ 
\midrule
Ours (w/o CFG) & \textbf{30.7952} & \textbf{0.0471} & 10.0613 & \textbf{26.1088} & 0.1021 & 17.7244 & \textbf{21.4909} & 0.2380 & 42.5515 \\
Ours (w/ CFG) & 30.0785 & 0.0494 & \textbf{9.8904} & 25.2294 & \textbf{0.1006} & \textbf{16.2667} & 20.5896 & \textbf{0.2019} & \textbf{25.2748} \\
\bottomrule
\end{tabular}}
\label{tab:cfg}
\end{table}

\begin{figure}[ht]
    \centering
    \includegraphics[width=\linewidth]{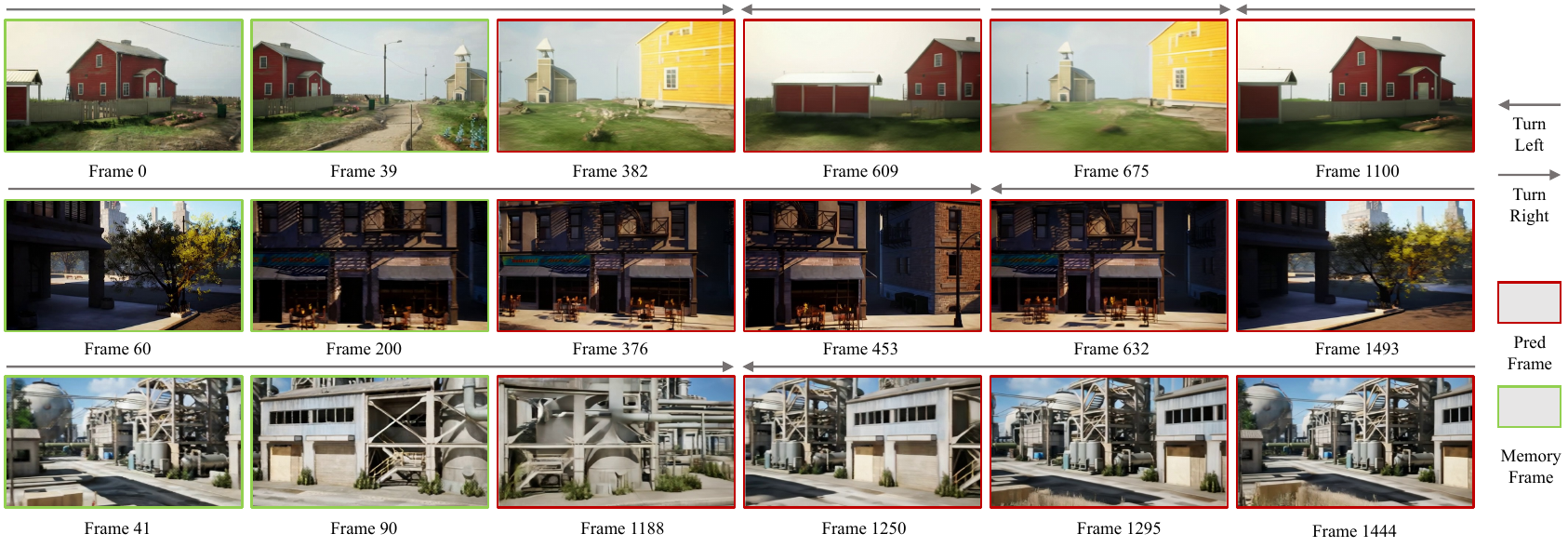}
    \caption{Long video generation on Context as Memory~\citep{yu2025contextas} dataset.}
    \label{fig:cam_data}
\end{figure}

\section{More Visualization Results}
\begin{figure}[!htbp]
    \centering
    \includegraphics[width=\linewidth]{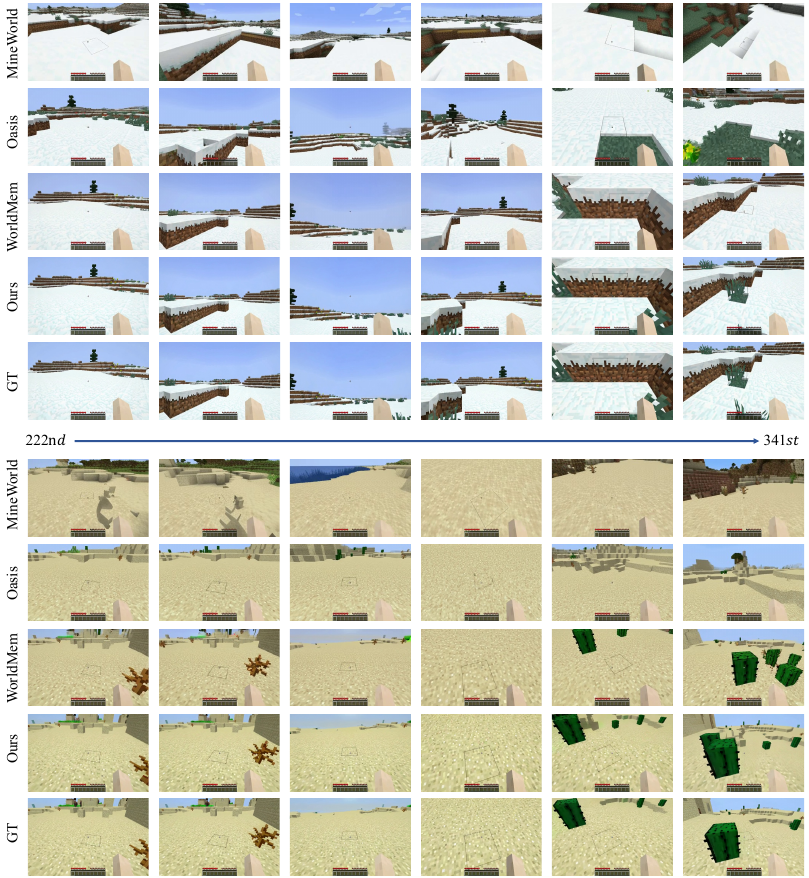}
    \caption{More qualitative comparison between our methods and other baselines.}
    \label{fig:com1}
\end{figure}

\begin{figure}[!htbp]
    \centering
    \includegraphics[width=\linewidth]{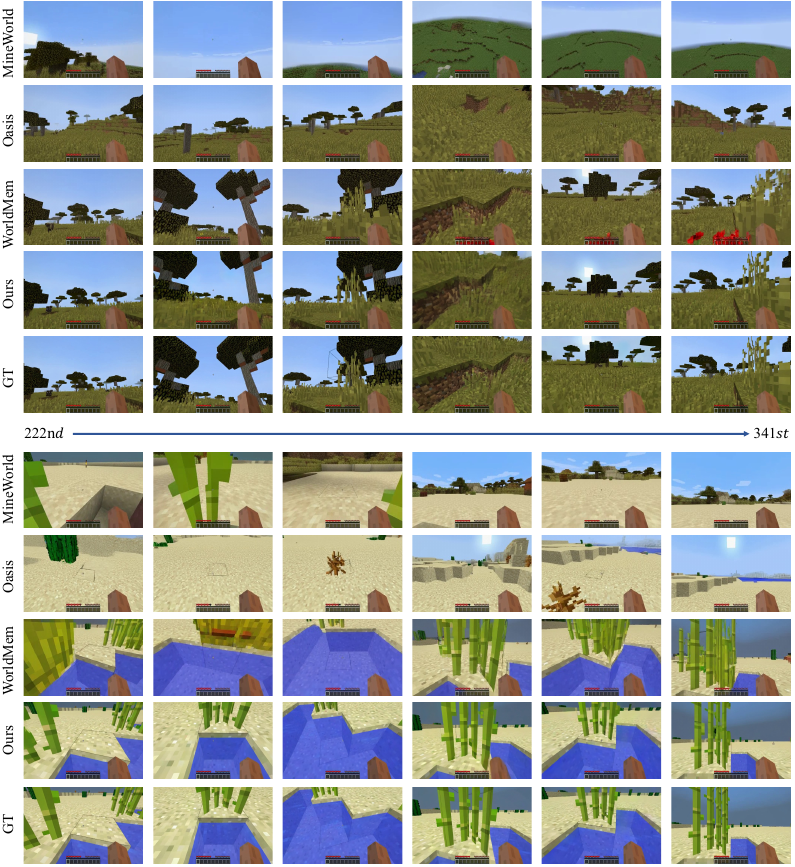}
    \caption{More qualitative comparison between our methods and other baselines.}
    \label{fig:com2}
\end{figure}

\begin{figure}[!t]
    \centering
    \includegraphics[width=\linewidth]{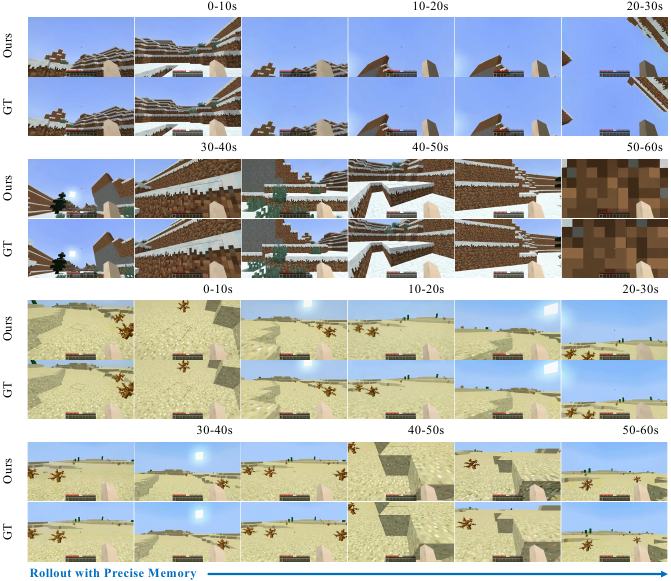}
    \caption{More visualization results of minute-long video generation.}
    \label{fig:more rollout}
\end{figure}

In~\cref{fig:com1} and~\cref{fig:com2}, following the setting in~\cref{qualitative exp}, we present additional comparisons with baseline methods. Across diverse scenarios, our model consistently demonstrates superior memory performance. Furthermore, as demonstrated in~\cref{fig:more rollout}, it is capable of inference for up to one minute while maintaining high fidelity. 

To demonstrate the effectiveness of our method across diverse datasets, we adopt the Context-as-Memory~\citep{yu2025contextas} dataset for both training and evaluation. This dataset contains abundant revisiting scenarios and can be used to assess the model’s memory capability. We drive the camera through a revisitation trajectory—repeated leftward and rightward pans—in three stylistically distinct environments: an island, a city, and a chemical plant, systematically probing the model's fine-grained spatiotemporal consistency upon re-entering previously visited regions. As shown in the~\cref{fig:cam_data}, our model faithfully reproduces previously observed structural layouts and local details across all three settings, demonstrating that the proposed memory mechanism sustains robust long-term consistency across diverse environments.

\section{Licenses}
\label{sec:license}
The WorldMem~\citep{xiao2025worldmem} datasets and code used in the main experiments is released under the S-Lab License 1.0. The baseline code bases including oasis~\citep{oasis2024}, MineWorld~\citep{guo2025mineworld} and Matrix-Game~\citep{he2025matrix-game-2} are all released under the MIT License. HY-WorldPlay~\citep{sun2025worldplay} is released under the TENCENT HY-WORLDPLAY COMMUNITY LICENSE AGREEMENT. We have strictly adhered to the terms and usage conditions of all the aforementioned licenses throughout our experiments.

\section{Broader Impacts and Limitations}
\label{sec:impact_lim}
\textbf{Broader Impacts}
The method proposed in this work aims to improve the spatiotemporal consistency of world models and to enhance their long-horizon extrapolation capability. It can be applied to some applications including gaming, virtual simulation, embodied AI, and film creation. At the same time, as a controllable approach capable of synthesizing long-duration, highly consistent videos, our work may inadvertently amplify the risks of technological misuse. Specifically, the ability to generate temporally extended and spatiotemporally coherent video could be exploited for fraudulent forgery and may substantially lower the barrier to producing disinformation at scale. We therefore call upon the community to strengthen defensive research directions—such as forgery detection and content provenance tracing—as essential mitigation measures against these risks.

\textbf{Limitations}
Our research focuses primarily on solving the precise memory and extrapolation generalization rather than inference acceleration via distillation, so real-time performance has not yet been achieved. In the future work, we will focus on developing an efficient real-time world model with hybrid memory mechanisms combining compressed global memory and fine-grained object-level memory, further improving the long-term consistency.

%% file: neurips_2026.bbl
\begin{thebibliography}{57}
\providecommand{\natexlab}[1]{#1}
\providecommand{\url}[1]{\texttt{#1}}
\expandafter\ifx\csname urlstyle\endcsname\relax
  \providecommand{\doi}[1]{doi: #1}\else
  \providecommand{\doi}{doi: \begingroup \urlstyle{rm}\Url}\fi

\bibitem[Ball et~al.(2025)Ball, Bauer, Belletti, Brownfield, Ephrat, Fruchter, Gupta, Holsheimer, Holynski, Hron, Kaplanis, Limont, McGill, Oliveira, Parker-Holder, Perbet, Scully, Shar, Spencer, Tov, Villegas, Wang, Yung, Baetu, Berbel, Bridson, Bruce, Buttimore, Chakera, Chandra, Collins, Cullum, Damoc, Dasagi, Gazeau, Gbadamosi, Han, Hirst, Kachra, Kerley, Kjems, Knoepfel, Koriakin, Lo, Lu, Mehring, Moufarek, Nandwani, Oliveira, Pardo, Park, Pierson, Poole, Ran, Salimans, Sanchez, Saprykin, Shen, Sidhwani, Smith, Stanton, Tomlinson, Vijaykumar, Wang, Wingfield, Wong, Xu, Yew, Young, Zubov, Eck, Erhan, Kavukcuoglu, Hassabis, Gharamani, Hadsell, van~den Oord, Mosseri, Bolton, Singh, and Rockt{\"a}schel]{genie3}
Philip~J. Ball, Jakob Bauer, Frank Belletti, Bethanie Brownfield, Ariel Ephrat, Shlomi Fruchter, Agrim Gupta, Kristian Holsheimer, Aleksander Holynski, Jiri Hron, Christos Kaplanis, Marjorie Limont, Matt McGill, Yanko Oliveira, Jack Parker-Holder, Frank Perbet, Guy Scully, Jeremy Shar, Stephen Spencer, Omer Tov, Ruben Villegas, Emma Wang, Jessica Yung, Cip Baetu, Jordi Berbel, David Bridson, Jake Bruce, Gavin Buttimore, Sarah Chakera, Bilva Chandra, Paul Collins, Alex Cullum, Bogdan Damoc, Vibha Dasagi, Maxime Gazeau, Charles Gbadamosi, Woohyun Han, Ed~Hirst, Ashyana Kachra, Lucie Kerley, Kristian Kjems, Eva Knoepfel, Vika Koriakin, Jessica Lo, Cong Lu, Zeb Mehring, Alex Moufarek, Henna Nandwani, Valeria Oliveira, Fabio Pardo, Jane Park, Andrew Pierson, Ben Poole, Helen Ran, Tim Salimans, Manuel Sanchez, Igor Saprykin, Amy Shen, Sailesh Sidhwani, Duncan Smith, Joe Stanton, Hamish Tomlinson, Dimple Vijaykumar, Luyu Wang, Piers Wingfield, Nat Wong, Keyang Xu, Christopher Yew, Nick Young, Vadim Zubov, Douglas
  Eck, Dumitru Erhan, Koray Kavukcuoglu, Demis Hassabis, Zoubin Gharamani, Raia Hadsell, A{\"a}ron van~den Oord, Inbar Mosseri, Adrian Bolton, Satinder Singh, and Tim Rockt{\"a}schel.
\newblock Genie 3: A new frontier for world models.
\newblock 2025.

\bibitem[Chen et~al.(2024)Chen, Mart{\'\i}~Mons{\'o}, Du, Simchowitz, Tedrake, and Sitzmann]{chen2024diffusion_forcing}
Boyuan Chen, Diego Mart{\'\i}~Mons{\'o}, Yilun Du, Max Simchowitz, Russ Tedrake, and Vincent Sitzmann.
\newblock Diffusion forcing: Next-token prediction meets full-sequence diffusion.
\newblock \emph{Advances in Neural Information Processing Systems}, 37:\penalty0 24081--24125, 2024.

\bibitem[Chen et~al.(2025{\natexlab{a}})Chen, Lin, Yang, Lin, Zhu, Fan, Zhang, Chen, Chen, Ma, et~al.]{chen2025skyreels}
Guibin Chen, Dixuan Lin, Jiangping Yang, Chunze Lin, Junchen Zhu, Mingyuan Fan, Hao Zhang, Sheng Chen, Zheng Chen, Chengcheng Ma, et~al.
\newblock Skyreels-v2: Infinite-length film generative model.
\newblock \emph{arXiv preprint arXiv:2504.13074}, 2025{\natexlab{a}}.

\bibitem[Chen et~al.(2026)Chen, Liang, Zhou, Ding, Liu, Wan, and Bai]{chen2026out}
Kaijin Chen, Dingkang Liang, Xin Zhou, Yikang Ding, Xiaoqiang Liu, Pengfei Wan, and Xiang Bai.
\newblock Out of sight but not out of mind: Hybrid memory for dynamic video world models.
\newblock \emph{arXiv preprint arXiv:2603.25716}, 2026.

\bibitem[Chen et~al.(2025{\natexlab{b}})Chen, Hu, Ding, and Jin]{chen2025vrag}
Taiye Chen, Xun Hu, Zihan Ding, and Chi Jin.
\newblock Learning world models for interactive video generation.
\newblock \emph{arXiv preprint arXiv:2505.21996}, 2025{\natexlab{b}}.

\bibitem[Cui et~al.(2026)Cui, Wu, Li, Yang, Li, Wang, Bai, Ban, and Hsieh]{cui2026lol}
Justin Cui, Jie Wu, Ming Li, Tao Yang, Xiaojie Li, Rui Wang, Andrew Bai, Yuanhao Ban, and Cho-Jui Hsieh.
\newblock Lol: Longer than longer, scaling video generation to hour.
\newblock \emph{arXiv preprint arXiv:2601.16914}, 2026.

\bibitem[Decart et~al.(2024)Decart, Quevedo, McIntyre, Campbell, Chen, and Wachen]{oasis2024}
Decart, Julian Quevedo, Quinn McIntyre, Spruce Campbell, Xinlei Chen, and Robert Wachen.
\newblock Oasis: A universe in a transformer.
\newblock 2024.
\newblock URL \url{https://oasis-model.github.io/}.
\newblock Project website.

\bibitem[Duan et~al.(2026)Duan, Xia, Zhang, Zhang, Zhou, Gou, He, Chen, Zhang, and Liu]{duan2026liveworld}
Zicheng Duan, Jiatong Xia, Zeyu Zhang, Wenbo Zhang, Gengze Zhou, Chenhui Gou, Yefei He, Feng Chen, Xinyu Zhang, and Lingqiao Liu.
\newblock Liveworld: Simulating out-of-sight dynamics in generative video world models.
\newblock \emph{arXiv preprint arXiv:2603.07145}, 2026.

\bibitem[Feng et~al.(2024)Feng, Zhang, Yang, Xiao, Shu, Liu, Zheng, Huang, Liu, and Zhang]{feng2024thematrix}
Ruili Feng, Han Zhang, Zhantao Yang, Jie Xiao, Zhilei Shu, Zhiheng Liu, Andy Zheng, Yukun Huang, Yu~Liu, and Hongyang Zhang.
\newblock The matrix: Infinite-horizon world generation with real-time moving control.
\newblock \emph{arXiv preprint arXiv:2412.03568}, 2024.

\bibitem[Guo et~al.(2025)Guo, Ye, He, Wu, Jiang, Pearce, and Bian]{guo2025mineworld}
Junliang Guo, Yang Ye, Tianyu He, Haoyu Wu, Yushu Jiang, Tim Pearce, and Jiang Bian.
\newblock Mineworld: a real-time and open-source interactive world model on minecraft.
\newblock \emph{arXiv preprint arXiv:2504.08388}, 2025.

\bibitem[He et~al.(2025)He, Peng, Liu, Wang, Zhang, Cui, Kang, Jiang, An, Ren, et~al.]{he2025matrix-game-2}
Xianglong He, Chunli Peng, Zexiang Liu, Boyang Wang, Yifan Zhang, Qi~Cui, Fei Kang, Biao Jiang, Mengyin An, Yangyang Ren, et~al.
\newblock Matrix-game 2.0: An open-source real-time and streaming interactive world model.
\newblock \emph{arXiv preprint arXiv:2508.13009}, 2025.

\bibitem[Henschel et~al.(2025)Henschel, Khachatryan, Poghosyan, Hayrapetyan, Tadevosyan, Wang, Navasardyan, and Shi]{henschel2025streamingt2v}
Roberto Henschel, Levon Khachatryan, Hayk Poghosyan, Daniil Hayrapetyan, Vahram Tadevosyan, Zhangyang Wang, Shant Navasardyan, and Humphrey Shi.
\newblock Streamingt2v: Consistent, dynamic, and extendable long video generation from text.
\newblock In \emph{Proceedings of the Computer Vision and Pattern Recognition Conference}, pages 2568--2577, 2025.

\bibitem[Heusel et~al.(2017)Heusel, Ramsauer, Unterthiner, Nessler, and Hochreiter]{heusel2017rfid}
Martin Heusel, Hubert Ramsauer, Thomas Unterthiner, Bernhard Nessler, and Sepp Hochreiter.
\newblock Gans trained by a two time-scale update rule converge to a local nash equilibrium.
\newblock \emph{Advances in neural information processing systems}, 30, 2017.

\bibitem[Hong et~al.(2025)Hong, Mei, Ge, Xu, Zhou, Bi, Hold-Geoffroy, Roberts, Fisher, Shechtman, et~al.]{hong2025relic}
Yicong Hong, Yiqun Mei, Chongjian Ge, Yiran Xu, Yang Zhou, Sai Bi, Yannick Hold-Geoffroy, Mike Roberts, Matthew Fisher, Eli Shechtman, et~al.
\newblock Relic: Interactive video world model with long-horizon memory.
\newblock \emph{arXiv preprint arXiv:2512.04040}, 2025.

\bibitem[Huang et~al.(2025{\natexlab{a}})Huang, Hu, Han, Shi, Tian, He, and Jiang]{huang2025memory_forcing}
Junchao Huang, Xinting Hu, Boyao Han, Shaoshuai Shi, Zhuotao Tian, Tianyu He, and Li~Jiang.
\newblock Memory forcing: Spatio-temporal memory for consistent scene generation on minecraft.
\newblock \emph{arXiv preprint arXiv:2510.03198}, 2025{\natexlab{a}}.

\bibitem[Huang et~al.(2025{\natexlab{b}})Huang, Li, He, Zhou, and Shechtman]{huang2025self_forcing}
Xun Huang, Zhengqi Li, Guande He, Mingyuan Zhou, and Eli Shechtman.
\newblock Self forcing: Bridging the train-test gap in autoregressive video diffusion.
\newblock \emph{arXiv preprint arXiv:2506.08009}, 2025{\natexlab{b}}.

\bibitem[Kim et~al.(2024)Kim, Kang, Choi, and Han]{kim2024fifo-diffuion}
Jihwan Kim, Junoh Kang, Jinyoung Choi, and Bohyung Han.
\newblock Fifo-diffusion: Generating infinite videos from text without training.
\newblock \emph{Advances in Neural Information Processing Systems}, 37:\penalty0 89834--89868, 2024.

\bibitem[Kondratyuk et~al.(2023)Kondratyuk, Yu, Gu, Lezama, Huang, Schindler, Hornung, Birodkar, Yan, Chiu, et~al.]{kondratyuk2023videopoet}
Dan Kondratyuk, Lijun Yu, Xiuye Gu, Jos{\'e} Lezama, Jonathan Huang, Grant Schindler, Rachel Hornung, Vighnesh Birodkar, Jimmy Yan, Ming-Chang Chiu, et~al.
\newblock Videopoet: A large language model for zero-shot video generation.
\newblock \emph{arXiv preprint arXiv:2312.14125}, 2023.

\bibitem[Kong et~al.(2024)Kong, Tian, Zhang, Min, Dai, Zhou, Xiong, Li, Wu, Zhang, et~al.]{kong2024hunyuanvideo}
Weijie Kong, Qi~Tian, Zijian Zhang, Rox Min, Zuozhuo Dai, Jin Zhou, Jiangfeng Xiong, Xin Li, Bo~Wu, Jianwei Zhang, et~al.
\newblock Hunyuanvideo: A systematic framework for large video generative models.
\newblock \emph{arXiv preprint arXiv:2412.03603}, 2024.

\bibitem[Li et~al.(2026)Li, Liu, Lin, and Chandraker]{li2026rollingsink}
Haodong Li, Shaoteng Liu, Zhe Lin, and Manmohan Chandraker.
\newblock Rolling sink: Bridging limited-horizon training and open-ended testing in autoregressive video diffusion.
\newblock \emph{arXiv preprint arXiv:2602.07775}, 2026.

\bibitem[Li et~al.(2025{\natexlab{a}})Li, Tang, Xu, Wu, Zhou, Shao, Yu, Cao, and Lu]{li2025hunyuan-gamecraft}
Jiaqi Li, Junshu Tang, Zhiyong Xu, Longhuang Wu, Yuan Zhou, Shuai Shao, Tianbao Yu, Zhiguo Cao, and Qinglin Lu.
\newblock Hunyuan-gamecraft: High-dynamic interactive game video generation with hybrid history condition.
\newblock \emph{arXiv preprint arXiv:2506.17201}, 2025{\natexlab{a}}.

\bibitem[Li et~al.(2025{\natexlab{b}})Li, Yi, Liu, Gao, Ma, and Kanazawa]{li2025cameras_prope}
Ruilong Li, Brent Yi, Junchen Liu, Hang Gao, Yi~Ma, and Angjoo Kanazawa.
\newblock Cameras as relative positional encoding.
\newblock \emph{arXiv preprint arXiv:2507.10496}, 2025{\natexlab{b}}.

\bibitem[Li et~al.(2025{\natexlab{c}})Li, Torr, Vedaldi, and Jakab]{li2025vmem}
Runjia Li, Philip Torr, Andrea Vedaldi, and Tomas Jakab.
\newblock Vmem: Consistent interactive video scene generation with surfel-indexed view memory.
\newblock \emph{arXiv preprint arXiv:2506.18903}, 2025{\natexlab{c}}.

\bibitem[Liu et~al.(2025)Liu, Hu, Xu, Shan, and Lu]{liu2025rollingforcing}
Kunhao Liu, Wenbo Hu, Jiale Xu, Ying Shan, and Shijian Lu.
\newblock Rolling forcing: Autoregressive long video diffusion in real time.
\newblock \emph{arXiv preprint arXiv:2509.25161}, 2025.

\bibitem[Liu et~al.(2022)Liu, Gong, and Liu]{liu2022rectiflow}
Xingchao Liu, Chengyue Gong, and Qiang Liu.
\newblock Flow straight and fast: Learning to generate and transfer data with rectified flow.
\newblock \emph{arXiv preprint arXiv:2209.03003}, 2022.

\bibitem[Lu et~al.(2024)Lu, Liang, Zhu, and Yang]{lu2024freelong}
Yu~Lu, Yuanzhi Liang, Linchao Zhu, and Yi~Yang.
\newblock Freelong: Training-free long video generation with spectralblend temporal attention.
\newblock \emph{Advances in Neural Information Processing Systems}, 37:\penalty0 131434--131455, 2024.

\bibitem[Mao et~al.(2025)Mao, Li, Li, Xu, Ying, He, Pang, Qiao, and Zhang]{mao2025yume-1.5}
Xiaofeng Mao, Zhen Li, Chuanhao Li, Xiaojie Xu, Kaining Ying, Tong He, Jiangmiao Pang, Yu~Qiao, and Kaipeng Zhang.
\newblock Yume-1.5: A text-controlled interactive world generation model.
\newblock \emph{arXiv preprint arXiv:2512.22096}, 2025.

\bibitem[Miyato et~al.(2023)Miyato, Jaeger, Welling, and Geiger]{miyato2023gta_rope}
Takeru Miyato, Bernhard Jaeger, Max Welling, and Andreas Geiger.
\newblock Gta: A geometry-aware attention mechanism for multi-view transformers.
\newblock \emph{arXiv preprint arXiv:2310.10375}, 2023.

\bibitem[Peebles and Xie(2023)]{peebles2023dit}
William Peebles and Saining Xie.
\newblock Scalable diffusion models with transformers.
\newblock In \emph{Proceedings of the IEEE/CVF international conference on computer vision}, pages 4195--4205, 2023.

\bibitem[Qiu et~al.(2023)Qiu, Xia, Zhang, He, Wang, Shan, and Liu]{qiu2023freenoise}
Haonan Qiu, Menghan Xia, Yong Zhang, Yingqing He, Xintao Wang, Ying Shan, and Ziwei Liu.
\newblock Freenoise: Tuning-free longer video diffusion via noise rescheduling.
\newblock \emph{arXiv preprint arXiv:2310.15169}, 2023.

\bibitem[Rombach et~al.(2022)Rombach, Blattmann, Lorenz, Esser, and Ommer]{rombach2022stablediffusoon}
Robin Rombach, Andreas Blattmann, Dominik Lorenz, Patrick Esser, and Bj{\"o}rn Ommer.
\newblock High-resolution image synthesis with latent diffusion models.
\newblock In \emph{Proceedings of the IEEE/CVF conference on computer vision and pattern recognition}, pages 10684--10695, 2022.

\bibitem[Shin et~al.(2025)Shin, Li, Zhang, Zhu, Park, Shechtman, and Huang]{shin2025motionstream}
Joonghyuk Shin, Zhengqi Li, Richard Zhang, Jun-Yan Zhu, Jaesik Park, Eli Shechtman, and Xun Huang.
\newblock Motionstream: Real-time video generation with interactive motion controls.
\newblock \emph{arXiv preprint arXiv:2511.01266}, 2025.

\bibitem[Su et~al.(2024)Su, Ahmed, Lu, Pan, Bo, and Liu]{su2024rope}
Jianlin Su, Murtadha Ahmed, Yu~Lu, Shengfeng Pan, Wen Bo, and Yunfeng Liu.
\newblock Roformer: Enhanced transformer with rotary position embedding.
\newblock \emph{Neurocomputing}, 568:\penalty0 127063, 2024.

\bibitem[Sun et~al.(2025)Sun, Zhang, Wang, Wu, Wang, Wang, Wang, Zhang, Wang, and Guo]{sun2025worldplay}
Wenqiang Sun, Haiyu Zhang, Haoyuan Wang, Junta Wu, Zehan Wang, Zhenwei Wang, Yunhong Wang, Jun Zhang, Tengfei Wang, and Chunchao Guo.
\newblock Worldplay: Towards long-term geometric consistency for real-time interactive world modeling.
\newblock \emph{arXiv preprint arXiv:2512.14614}, 2025.

\bibitem[Tang et~al.(2025)Tang, Liu, Li, Wu, Yang, Zhao, Gong, Yuan, Shao, and Lu]{tang2025hunyuan-gamecraft2}
Junshu Tang, Jiacheng Liu, Jiaqi Li, Longhuang Wu, Haoyu Yang, Penghao Zhao, Siruis Gong, Xiang Yuan, Shuai Shao, and Qinglin Lu.
\newblock Hunyuan-gamecraft-2: Instruction-following interactive game world model.
\newblock \emph{arXiv preprint arXiv:2511.23429}, 2025.

\bibitem[Team et~al.(2025)Team, Wang, Liu, Wu, Gu, Wang, Zuo, Huang, Li, Zhang, et~al.]{team2025hunyuanworld}
HunyuanWorld Team, Zhenwei Wang, Yuhao Liu, Junta Wu, Zixiao Gu, Haoyuan Wang, Xuhui Zuo, Tianyu Huang, Wenhuan Li, Sheng Zhang, et~al.
\newblock Hunyuanworld 1.0: Generating immersive, explorable, and interactive 3d worlds from words or pixels.
\newblock \emph{arXiv preprint arXiv:2507.21809}, 2025.

\bibitem[Teng et~al.(2025)Teng, Jia, Sun, Li, Li, Tang, Han, Zhang, Zhang, Luo, et~al.]{teng2025magi1}
Hansi Teng, Hongyu Jia, Lei Sun, Lingzhi Li, Maolin Li, Mingqiu Tang, Shuai Han, Tianning Zhang, WQ~Zhang, Weifeng Luo, et~al.
\newblock Magi-1: Autoregressive video generation at scale.
\newblock \emph{arXiv preprint arXiv:2505.13211}, 2025.

\bibitem[Valevski et~al.(2024)Valevski, Leviathan, Arar, and Fruchter]{valevski2024GameNgen}
Dani Valevski, Yaniv Leviathan, Moab Arar, and Shlomi Fruchter.
\newblock Diffusion models are real-time game engines.
\newblock \emph{arXiv preprint arXiv:2408.14837}, 2024.

\bibitem[Wan et~al.(2025)Wan, Wang, Ai, Wen, Mao, Xie, Chen, Yu, Zhao, Yang, et~al.]{wan2025wan}
Team Wan, Ang Wang, Baole Ai, Bin Wen, Chaojie Mao, Chen-Wei Xie, Di~Chen, Feiwu Yu, Haiming Zhao, Jianxiao Yang, et~al.
\newblock Wan: Open and advanced large-scale video generative models.
\newblock \emph{arXiv preprint arXiv:2503.20314}, 2025.

\bibitem[Wang et~al.(2026)Wang, Lin, Yoon, Cho, Zhang, and Bansal]{wang2026anchorweave}
Zun Wang, Han Lin, Jaehong Yoon, Jaemin Cho, Yue Zhang, and Mohit Bansal.
\newblock Anchorweave: World-consistent video generation with retrieved local spatial memories.
\newblock \emph{arXiv preprint arXiv:2602.14941}, 2026.

\bibitem[Wu et~al.(2025)Wu, Yang, Po, Xu, Liu, Lin, and Wetzstein]{wu2025spmem}
Tong Wu, Shuai Yang, Ryan Po, Yinghao Xu, Ziwei Liu, Dahua Lin, and Gordon Wetzstein.
\newblock Video world models with long-term spatial memory.
\newblock \emph{arXiv preprint arXiv:2506.05284}, 2025.

\bibitem[Xiang et~al.(2026)Xiang, Liu, Zhang, Yang, Fang, Wang, Wang, Zou, Su, and Zhu]{xiang2026viewrope}
Chendong Xiang, Jiajun Liu, Jintao Zhang, Xiao Yang, Zhengwei Fang, Shizun Wang, Zijun Wang, Yingtian Zou, Hang Su, and Jun Zhu.
\newblock Geometry-aware rotary position embedding for consistent video world model.
\newblock \emph{arXiv preprint arXiv:2602.07854}, 2026.

\bibitem[Xiang et~al.(2025)Xiang, Gu, Liu, Feng, Gao, Hu, Huang, Liu, Yang, Zhou, et~al.]{xiang2025pan}
Jiannan Xiang, Yi~Gu, Zihan Liu, Zeyu Feng, Qiyue Gao, Yiyan Hu, Benhao Huang, Guangyi Liu, Yichi Yang, Kun Zhou, et~al.
\newblock Pan: A world model for general, interactable, and long-horizon world simulation.
\newblock \emph{arXiv preprint arXiv:2511.09057}, 2025.

\bibitem[Xiao et~al.(2025)Xiao, Lan, Zhou, Ouyang, Yang, Zeng, and Pan]{xiao2025worldmem}
Zeqi Xiao, Yushi Lan, Yifan Zhou, Wenqi Ouyang, Shuai Yang, Yanhong Zeng, and Xingang Pan.
\newblock Worldmem: Long-term consistent world simulation with memory.
\newblock \emph{arXiv preprint arXiv:2504.12369}, 2025.

\bibitem[Yang et~al.(2025)Yang, Huang, Chu, Xiao, Zhao, Wang, Li, Xie, Chen, Lu, et~al.]{yang2025longlive}
Shuai Yang, Wei Huang, Ruihang Chu, Yicheng Xiao, Yuyang Zhao, Xianbang Wang, Muyang Li, Enze Xie, Yingcong Chen, Yao Lu, et~al.
\newblock Longlive: Real-time interactive long video generation.
\newblock \emph{arXiv preprint arXiv:2509.22622}, 2025.

\bibitem[Yang et~al.(2024)Yang, Teng, Zheng, Ding, Huang, Xu, Yang, Hong, Zhang, Feng, et~al.]{yang2024cogvideox}
Zhuoyi Yang, Jiayan Teng, Wendi Zheng, Ming Ding, Shiyu Huang, Jiazheng Xu, Yuanming Yang, Wenyi Hong, Xiaohan Zhang, Guanyu Feng, et~al.
\newblock Cogvideox: Text-to-video diffusion models with an expert transformer.
\newblock \emph{arXiv preprint arXiv:2408.06072}, 2024.

\bibitem[Ye et~al.(2025)Ye, Zhou, Lv, Ma, Zhang, Lv, Li, Deng, Yang, Fu, et~al.]{ye2025yan}
Deheng Ye, Fangyun Zhou, Jiacheng Lv, Jianqi Ma, Jun Zhang, Junyan Lv, Junyou Li, Minwen Deng, Mingyu Yang, Qiang Fu, et~al.
\newblock Yan: Foundational interactive video generation.
\newblock \emph{arXiv preprint arXiv:2508.08601}, 2025.

\bibitem[Yi et~al.(2025)Yi, Jang, Cho, Nam, Yoon, and Kim]{yi2025deepforcing}
Jung Yi, Wooseok Jang, Paul~Hyunbin Cho, Jisu Nam, Heeji Yoon, and Seungryong Kim.
\newblock Deep forcing: Training-free long video generation with deep sink and participative compression.
\newblock \emph{arXiv preprint arXiv:2512.05081}, 2025.

\bibitem[Yin et~al.(2025)Yin, Zhang, Zhang, Freeman, Durand, Shechtman, and Huang]{yin2025causvid}
Tianwei Yin, Qiang Zhang, Richard Zhang, William~T Freeman, Fredo Durand, Eli Shechtman, and Xun Huang.
\newblock From slow bidirectional to fast autoregressive video diffusion models.
\newblock In \emph{Proceedings of the Computer Vision and Pattern Recognition Conference}, pages 22963--22974, 2025.

\bibitem[Yu et~al.(2025{\natexlab{a}})Yu, Bai, Qin, Liu, Wang, Wan, Zhang, and Liu]{yu2025contextas}
Jiwen Yu, Jianhong Bai, Yiran Qin, Quande Liu, Xintao Wang, Pengfei Wan, Di~Zhang, and Xihui Liu.
\newblock Context as memory: Scene-consistent interactive long video generation with memory retrieval.
\newblock \emph{arXiv preprint arXiv:2506.03141}, 2025{\natexlab{a}}.

\bibitem[Yu et~al.(2025{\natexlab{b}})Yu, Qin, Wang, Wan, Zhang, and Liu]{yu2025gamefactory}
Jiwen Yu, Yiran Qin, Xintao Wang, Pengfei Wan, Di~Zhang, and Xihui Liu.
\newblock Gamefactory: Creating new games with generative interactive videos.
\newblock \emph{arXiv preprint arXiv:2501.08325}, 2025{\natexlab{b}}.

\bibitem[Zhang and Sennrich(2019)]{zhang2019rmsnorm}
Biao Zhang and Rico Sennrich.
\newblock Root mean square layer normalization.
\newblock \emph{Advances in neural information processing systems}, 32, 2019.

\bibitem[Zhang et~al.(2018)Zhang, Isola, Efros, Shechtman, and Wang]{zhang2018lpips}
Richard Zhang, Phillip Isola, Alexei~A Efros, Eli Shechtman, and Oliver Wang.
\newblock The unreasonable effectiveness of deep features as a perceptual metric.
\newblock In \emph{Proceedings of the IEEE conference on computer vision and pattern recognition}, pages 586--595, 2018.

\bibitem[Zhang et~al.(2025)Zhang, Peng, Wang, Wang, Zhu, Kang, Jiang, Gao, Li, Liu, et~al.]{zhang2025matrix-game}
Yifan Zhang, Chunli Peng, Boyang Wang, Puyi Wang, Qingcheng Zhu, Fei Kang, Biao Jiang, Zedong Gao, Eric Li, Yang Liu, et~al.
\newblock Matrix-game: Interactive world foundation model.
\newblock \emph{arXiv preprint arXiv:2506.18701}, 2025.

\bibitem[Zhao et~al.(2025{\natexlab{a}})Zhao, Wei, Liu, Zhang, Xu, and Lu]{zhao2025spatia}
Jinjing Zhao, Fangyun Wei, Zhening Liu, Hongyang Zhang, Chang Xu, and Yan Lu.
\newblock Spatia: Video generation with updatable spatial memory.
\newblock \emph{arXiv preprint arXiv:2512.15716}, 2025{\natexlab{a}}.

\bibitem[Zhao et~al.(2025{\natexlab{b}})Zhao, He, Chen, Zhu, Li, and Zhu]{zhao2025riflex}
Min Zhao, Guande He, Yixiao Chen, Hongzhou Zhu, Chongxuan Li, and Jun Zhu.
\newblock Riflex: A free lunch for length extrapolation in video diffusion transformers.
\newblock \emph{arXiv preprint arXiv:2502.15894}, 2025{\natexlab{b}}.

\bibitem[Zhao et~al.(2025{\natexlab{c}})Zhao, Zhu, Wang, Yan, Zhang, He, Yang, Li, and Zhu]{zhao2025ultravico}
Min Zhao, Hongzhou Zhu, Yingze Wang, Bokai Yan, Jintao Zhang, Guande He, Ling Yang, Chongxuan Li, and Jun Zhu.
\newblock Ultravico: Breaking extrapolation limits in video diffusion transformers.
\newblock \emph{arXiv preprint arXiv:2511.20123}, 2025{\natexlab{c}}.

\end{thebibliography}
